\pdfoutput=1
\documentclass{fairmeta}
\usepackage{amsmath}
\usepackage{pifont}
\usepackage{array}
\usepackage{siunitx}
\usepackage{float}
\usepackage{enumitem}
\usepackage{xurl}

\newcommand{\playerone}{{\sffamily\bfseries Player-1}}
\newcommand{\playerdata}{{\sffamily\bfseries Player-1-Data}}
\newcommand{\playergym}{{\sffamily\bfseries Player-1-Gym}}
\newcommand{\playernineb}{Player-1-9B}
\newcommand{\playerthirtyfiveb}{Player-1-35B-A3B}
\newcommand{\playeronesft}{Player-1-SFT}
\newcommand{\playerninebsft}{Player-1-9B-SFT}
\newcommand{\playerthirtyfivebsft}{Player-1-35B-A3B-SFT}
\newcommand{\papertablesize}{\fontsize{8}{9}\selectfont}

\definecolor{projectpagelink}{HTML}{DA2F8A}
\hypersetup{pdftitle={Player-1: Learning to Play from Synthetic Experience},
  pdfauthor={Mingyu Ouyang, Yiran Wang, Siyuan Hu, Xingyilang Yin, Huiqiang Sun, Guangzhi Wang, Junfu Pu, Cheng Cheng, Honghao Chen, Mike Zheng Shou, Ying Shan, Hwee Tou Ng}}

\begin{document}

\title{Player-1:\\ Learning to Play from Synthetic Experience}
\author[1,2,\ast]{Mingyu~Ouyang}
\author[2,\ddagger]{Yiran~Wang}
\author[3]{Siyuan~Hu}
\author[2]{Xingyilang~Yin}
\author[2]{Huiqiang~Sun}
\author[2]{Guangzhi~Wang}
\author[2]{Junfu~Pu}
\author[2]{Cheng~Cheng}
\author[2]{Honghao~Chen}
\author[1,\ddagger]{Mike~Zheng~Shou}
\author[2]{Ying~Shan}
\author[1,\ddagger]{Hwee~Tou~Ng}
\affiliation[1]{National University of Singapore}
\affiliation[2]{ARC Lab, Tencent IEG}
\affiliation[3]{Nanyang Technological University}
\metadata[Project]{\href{https://player-1-project.github.io/}{\textcolor{projectpagelink}{player-1-project.github.io/}}}
\metadatatwo[Code]{\href{https://github.com/TencentARC/Player-1}{\textcolor{projectpagelink}{github.com/TencentARC/Player-1}}}
\metadatatwo[Data]{\href{https://huggingface.co/datasets/yyyang/Player-1-Data}{\textcolor{projectpagelink}{huggingface.co/datasets/yyyang/Player-1-Data}}}
\contribution[\ast]{Work done during internship at ARC Lab, Tencent IEG.}
\contribution[\ddagger]{Co-corresponding authors.}

\abstract{
Computer-use agents have demonstrated strong capabilities in interpreting screens and executing keyboard and mouse actions.
However, games, with their visually rich worlds and diverse mechanics, remain challenging and offer a demanding next step toward general digital control.
Current game-agent learning pipelines often rely on large-scale demonstrations with limited coverage of dynamic game states, while scalable environments for interactive training remain difficult to obtain.
We introduce \playerone{}, a post-training framework for learning to play from synthetic experience.
We extend the computer-use action space with relative pointer motion, sustained inputs, and composite actions.
For supervised initialization, we construct \playerdata{}, 6,044 curated gameplay trajectories with annotated task instructions and per-step reasoning.
To support interactive reinforcement learning, we develop \playergym{}, sixteen synthetic 2D and 3D environments designed around recurring control demands shared across games.
Programmatically verifiable rewards and randomized observations and control dynamics support multi-environment reinforcement learning with PPO.
The resulting \playernineb{} and \playerthirtyfiveb{} improve MineDojo-300 success over their SFT policies by 27.8 and 17.7 percentage points, respectively.
\playernineb{} reaches 38.33\% success compared with 23.29\% for the strongest evaluated baseline, with further gains over SFT on ViZDoom-10 and GameWorld.
Further analyses show increasing use of composite actions during RL and confirm that cross-game transfer scales with synthetic environment diversity.
We demonstrate that learning from synthetic experience substantially improves gameplay performance beyond supervised initialization and enables transfer beyond the RL training suite, while broadly preserving base-model general capabilities.
}
\maketitle

\section{Introduction}

Foundation models have developed broad capabilities in visual understanding, reasoning, and instruction following \citep{openai2024gpt4o,anthropic2025claude4,gemini2025gemini25}.
Computer-use agents bring these capabilities to digital interfaces, interpreting screens and executing keyboard and mouse actions \citep{wang2025uitars2,bai2025qwen3vl,lu2026qwencua}.
Games offer a demanding next step toward general digital control: their visually complex environments and diverse mechanics require precise, sustained, and concurrent control under dynamic and rapidly changing visual feedback \citep{luo2026aigames}.

Recent game agents learn from large-scale gameplay demonstrations and interactive experience \citep{wang2025gametars,magne2026nitrogen,sima2025two,shi2026odysseus,zhang2026generalist}.
Yet demonstrations cover only a limited portion of the states an agent may encounter in open-ended tasks, while scalable environments for interactive training remain difficult to obtain.
Despite differences in appearance and rules, we find that many games share recurring control demands such as precise aiming, sustained movement, and coordinated inputs.
This motivates our central question:
\textbf{\emph{Can synthetic experience advance game-playing capabilities beyond supervised initialization, with gains that generalize to games outside the RL training suite?}}

Three challenges shape this approach.
First, gameplay datasets recorded with heterogeneous input conventions and temporal resolutions must be converted into a shared action space, and paired with task-wise instructions and step-wise reasoning.
Second, learning beyond demonstrations requires feedback on states reached by the policy’s own actions, but implementing programmatic rewards across diverse commercial games is difficult to scale.
Third, experience from environments with different control demands, dynamics, and reward scales must be incorporated into a shared policy that generalizes beyond the training suite.
These challenges motivate our supervised initialization from curated demonstrations, followed by learning in controllable synthetic environments with verifiable rewards.

\begin{figure}[!t]
\centering
\includegraphics[width=\linewidth]{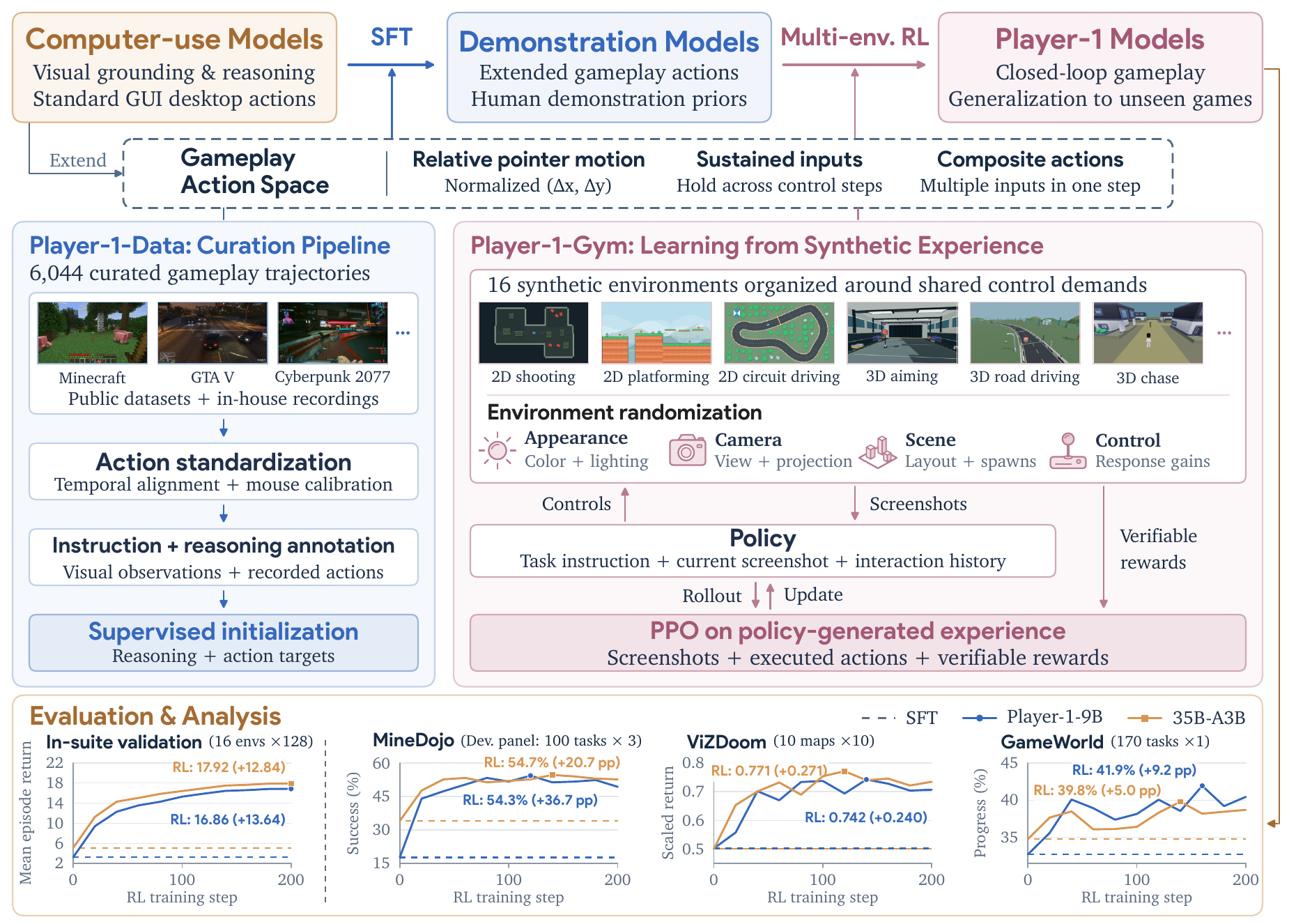}
\caption{Overview of \playerone{}: learning to play games beyond demonstrations.
A shared action space for gameplay connects supervised initialization on \playerdata{} with multi-environment RL in \playergym{}.
Learning from synthetic experience with verifiable rewards advances control capabilities beyond SFT, with gains that generalize beyond the RL training suite.}
\label{fig:overview}
\end{figure}

We present \playerone{}, a post-training framework for learning to play from synthetic experience, as illustrated in Figure~\ref{fig:overview}.
We extend the computer-use action space with relative pointer motion, sustained inputs, and composite actions.
\playerdata{} provides curated gameplay trajectories pairing recorded actions with annotated task instructions and step-wise reasoning for supervised initialization.
\playergym{} then provides sixteen synthetic 2D and 3D environments designed around recurring control demands across aiming, combat, driving, navigation, and UI interaction.
Programmatic game-state rewards and randomization of observations and control dynamics support multi-environment reinforcement learning with verifiable rewards (RLVR), yielding \playernineb{} and \playerthirtyfiveb{}.
Table~\ref{tab:game-agent-comparison} summarizes the differences and focuses of representative game-agent learning pipelines.
Across diverse benchmarks, learning from synthetic experience delivers substantial control gains that generalize well beyond the training environments, all while preserving the base model's general-purpose capabilities.

We summarize our main contributions as follows:
\begingroup
\setlength{\leftmargini}{1.5pc}
\begin{itemize}[itemsep=1pt,topsep=2pt]

\item
We present \playerone{}, a post-training framework for learning transferable gameplay capabilities from synthetic experience. It combines an extended action space for gameplay with \playerdata{}, a collection of 6,044 curated gameplay trajectories for supervised initialization.

\item
We introduce \playergym{}, sixteen synthetic 2D and 3D environments with verifiable rewards and randomization across four complementary dimensions for multi-environment reinforcement learning.

\item
We show that synthetic experience improves gameplay performance substantially beyond demonstrations, boosting MineDojo-300 success from 10.5\% to 38.3\% ($\sim 3.64\times$ over SFT, vs.\ 23.3\% for the top baseline) and generalizing across ViZDoom-10, GameWorld, and commercial games, while broadly preserving the base model's general-purpose capabilities.

\end{itemize}
\endgroup

\begin{table}[tbp]
\captionsetup{skip=1pt}

\centering
\caption{Comparison of representative game-agent learning pipelines, including supervision, RL environments, evaluation scope, and pipeline and model availability.}
\label{tab:game-agent-comparison}
\papertablesize
\setlength{\tabcolsep}{0.7pt}
\renewcommand{\arraystretch}{1.0}
\begin{tabular*}{\linewidth}{@{\extracolsep{\fill}}lllllc@{}}
\toprule
\textbf{Method} & \textbf{Learning focus} & \textbf{Supervision} & \textbf{RL environments} & \textbf{Evaluation} & \textbf{Availability} \\
\midrule
Game-TARS~\citep{wang2025gametars} & Large-scale pretraining
  & Reasoning + action
  & --- & Held-out games & \ding{55} \\
\addlinespace[0.5pt]
SIMA 2~\citep{sima2025two} & Embodied assistance
  & Reasoning + action
  & Games + world model & Held-out games & \ding{55} \\
\addlinespace[0.5pt]
NitroGen~\citep{magne2026nitrogen} & System-1 generalist
  & Action
  & --- & Games + adaptation & \ding{51} \\
\addlinespace[0.5pt]
Odysseus~\citep{shi2026odysseus} & Long-horizon RL
  & Reasoning + action
  & 5 Mario levels & Mario games & \ding{51} \\
\addlinespace[0.5pt]
GameScaling~\citep{zhang-scaling-games} & Continued pretraining
  & Reasoning + action
  & --- & Held-out games & \ding{51} \\
\midrule
\playerone{} & \textbf{Beyond demonstrations}
  & Reasoning + action
  & \textbf{16 synthetic envs} & \textbf{RL held-out games} & \ding{51}\\
\bottomrule
\end{tabular*}
\par\smallskip
\par

\vspace{-0.5em}

\end{table}

\section{Related Work}
\label{sec:related-work}

\paragraph{Generalist and Computer-Use Agents.}
Foundation models provide broad language, reasoning, and multimodal capabilities that underpin general-purpose agents \citep{openai2024gpt4o,anthropic2025claude4,gemini2025gemini25,yang2025qwen3}.
Vision-language models (VLMs) connect visual perception with language, supporting scene understanding and spatial grounding \citep{bai2025qwen3vl}.
Computer-use agents (CUAs) specialize these capabilities for desktop interaction, using keyboard and mouse actions to complete multi-step tasks \citep{hu2024dawn,qin2025uitars,wang2025uitars2,lu2026qwencua}.

\paragraph{Game-Playing and Game-Development Agents.}
Game-agent research largely focuses on gameplay, spanning planning-based systems \citep{tan2025cradle}, learned visual policies and vision--language--action (VLA) models \citep{sima2024,tan2025lumine,li2025jarvisvla,wang2025gametars,magne2026nitrogen,yue2026openp2p,guo2026gamewam,zhang-scaling-games}, and interactive reinforcement learning for self-improvement \citep{sima2025two,shi2026odysseus}.
Most recently, Jev explores low-latency game control through structured game states \citep{almeida2026jev}.
Complementary benchmarks evaluate gameplay, reflection, and decision making across diverse environments \citep{ouyang2026gameworld,zhang2026gameverse,hu2025lmgame,park2025orak,GameHorizonSuite2026}.
Beyond game-playing, a parallel line of work investigates game-development agents that autonomously design, code, and deploy games in real engines, evaluated by benchmarks such as GameDevBench \citep{chi2026gamedevbench} and GameCraft \citep{luo2026gamecraft}.

\paragraph{Agentic Post-Training with Verifiable Rewards.}
Post-training with verifiable rewards enables agents to learn from the outcomes of their own actions.
Environment-synthesis approaches expand training coverage by constructing interactive tasks with verifiable rewards \citep{dong2026agentworld,wang2026cuagym,tong2025gamerlsynthesizingmultimodalverifiable,khan2025dataenvgym}.
Adaptive approaches generate and adapt environments based on agent weaknesses or task curricula \citep{zala2024envgen,liu2026spade,wang2026rlanything}.
SIMA 2 explores self-improvement in Genie 3 environments using Gemini-based rewards \citep{sima2025two}.
\playerone{} combines supervised initialization with learning from synthetic experience in environments designed around shared control demands, demonstrating gains in gameplay performance that generalize beyond the RL training suite.

\section{Method}
\label{sec:method}

\playerone{} first extends computer-use actions to support dynamic gameplay, and initializes the policy on \playerdata{}.
We then optimize the policy across \playergym{}'s synthetic environments using reinforcement learning on verifiable rewards.

\subsection{Extending the Computer-Use Action Space}
\label{sec:game-control}
\label{sec:action-space}

We extend the computer-use action space for dynamic gameplay while retaining screenshot-conditioned decision making.
At turn $t$, the policy receives task instruction $q$, screenshot $I_t$, and interaction history $C_H(h_{<t})$ (see \hyperref[par:policy-context]{Section~\ref*{sec:player1-data}}):
\[
(z_t,\hat a_t)\sim\pi_\theta(\cdot\mid q,I_t,C_H(h_{<t})).
\]
Here $z_t$ is reasoning and $\hat a_t$ the proposed action.

\paragraph{Relative pointer motion.}
For mouse actions, CUAs typically use normalized absolute coordinates to select targets by specifying their on-screen positions.
In gameplay scenarios, to control the camera without a visible pointer, we extend this action space with relative displacements:
\[
\mathcal A_{\mathrm{pointer}}=
\underbrace{\{\operatorname{absolute}(x,y)\}}_{\text{Retained CUA coordinates}}
\;\cup\;
\underbrace{\text{\boldmath$\{\operatorname{relative}(\Delta x,\Delta y)\}$}}
_{\textbf{Relative-motion extension}}.
\]

Normalized coordinates $x,y\in[0,1000]$ and displacements $\Delta x,\Delta y\in[-1000,1000]$ map to screen pixels, with view rotation scaled by camera gain.

\paragraph{Sustained inputs.}
We extend taps and clicks with holds lasting one full control step of duration $\Delta t$.
For a key or mouse button $b$, the press or long-hold action space is
\[
\mathcal A_{\mathrm{press}}(b)=
\underbrace{\{\operatorname{tap/click}(b)\}}_{\text{Retained CUA taps/clicks}}
\;\cup\;
\underbrace{\text{\boldmath$\{\operatorname{hold}(b;\Delta t)\}$}}_{\textbf{Sustained extension}}.
\]
Repeating holds across consecutive control steps sustains movement and interaction.

\paragraph{Composite actions.}
Composite actions are defined as two or more compatible keyboard or mouse inputs executed concurrently within one control step.
The resulting step action space is formulated as:
\[
\mathcal A_{\mathrm{step}}=
\underbrace{\{\operatorname{wait}\}\cup\mathcal A_{\mathrm{keyboard}}
\cup\mathcal A_{\mathrm{mouse}}}_{\text{Individual CUA inputs}}
\;\cup\;
\underbrace{\text{\boldmath$\{\textbf{input}_{1}\parallel\cdots\parallel\textbf{input}_{m}\}$}}
_{\textbf{Composite-action extension}}.
\]
Each $\text{input}_i\in\mathcal A_{\mathrm{keyboard}}\cup\mathcal A_{\mathrm{mouse}}$; $\parallel$ denotes concurrent execution and $m\ge2$.
This enables concurrent motor coordination, such as camera-relative aiming and firing while holding directional movement.

\subsection{\playerdata: Supervised Initialization from Demonstrations}
\label{sec:supervised-initialization}
\label{sec:player1-data}

The extended action space defines the available controls, but the policy must learn when and how to execute them.
In this section, we present \playerdata{}, establishing a scalable data production pipeline alongside a curated demonstration dataset to initialize computer-use models for dynamic gameplay.

\begin{figure}[tbp]

\centering
\input{figures/figure2/overview_citation_links.tex}%

\vspace{-0.25em}

\caption{Overview of \playerdata{} for supervised initialization.
(a) Sources and scale, with sampled control hours. (b) Game distribution.
(c) Recorded controls summarized by control type, number of inputs per control step, and pointer-motion mode in our demonstration trajectories.
Two or more concurrent inputs form a composite action.
(d) A demonstration sample pairing recorded actions with an annotated task instruction and visual reasoning.}
\label{fig:p1-data-action-distribution}
\par

\vspace{-0.5em}

\end{figure}

\paragraph{Data production pipeline and dataset.}
To transform raw, heterogeneous human gameplay into standardized supervision, we design a three-stage production pipeline to construct \playerdata{}.
First, the ingestion stage curates 6,044 trajectories totaling 295,097 turns and 16.4 hours, spanning 172 diverse games across four public datasets \citep{choi2026d2e,markov2026gaming,yue2026openp2p,baker2022vpt} and in-house recordings (Figure~\ref{fig:p1-data-action-distribution}(a--b)).
Second, the standardization stage maps disparate inputs into our shared action space: timestamps are aligned to a 200-ms grid and mouse inputs are calibrated to screen displacements, yielding the composite actions and continuous pointer dynamics shown in Figure~\ref{fig:p1-data-action-distribution}(c).
Finally, the annotation stage uses a VLM \citep{deepmind2026gemini38flash} to synthesize task goals and per-turn visual reasoning conditioned on gameplay context (Figure~\ref{fig:p1-data-action-distribution}(d)).
Appendix~\ref{app:sft-details} details the pipeline implementation, calibration formulas, and annotation prompts.

\paragraph{Supervised fine-tuning.}
\label{par:policy-context}
We construct each SFT example as $(q,I_t,C_H(h_{<t}),z_t,a_t)$, where the task instruction $q$, current screenshot $I_t$, and retained history $C_H(h_{<t})$ form the policy input, while visual reasoning $z_t$ and action $a_t$ provide supervision. Following \citet{wang2025gametars}, our SFT, RL, and evaluation share a $2+2$ history setting: two recent visual--action turns and two earlier action-only turns, with reasoning omitted.
We fine-tune the policy to predict $(z_t,a_t)$ conditioned on the task instruction, current screenshot, and retained history. We use \emph{action-balanced sampling} detailed in Appendix~\ref{app:sft-training} to increase exposure to less frequent controls and composite actions.

\subsection{\playergym: Learning from Synthetic Experience}
\label{sec:player1-gym}
\label{sec:post-training}

We construct \playergym{}, a suite of synthetic environments organized around controls shared across games.
Programmatically verifiable rewards, together with randomized observations and control dynamics, support our multi-environment PPO training of the SFT-initialized policy.

\begin{figure}[tbp]
\centering
\includegraphics[width=\linewidth]{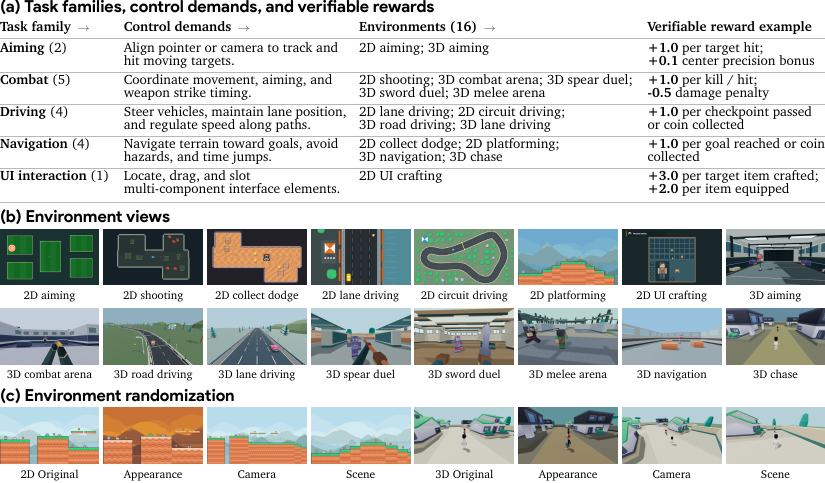}
\par
\vspace{-0.25em}
\caption{\playergym{}: synthetic experience with verifiable rewards.
(a) Five task families, their control demands, environments, and representative environment-specific rewards.
(b) Visualization of sixteen 2D and 3D environments.
(c) Examples of appearance, camera, and scene randomization applied.}
\label{fig:p1-gym-environments}

\vspace{-0.75em}

\end{figure}

\paragraph{Verifiable environment construction.}
We organize training tasks into five recurring control families: aiming, combat, driving, navigation, and UI interaction.
Across these families, we implement sixteen diverse 2D and 3D environments using shared engines and scene templates (Figure~\ref{fig:p1-gym-environments}(a--b)).
A unified runtime executes the shared action space while providing visual-only screenshot observations, keeping privileged internal states hidden from the policy.
To provide reliable reinforcement signals, each environment defines deterministic, task-specific scoring rules over underlying game states.
Score increments serve as dense, step-level verifiable rewards without relying on external learned judges.

\paragraph{Environment randomization.}
To further broaden training experience within each environment, we randomize four complementary dimensions in \playergym{}: \textbf{(1) Appearance}: colors, lighting, and assets; \textbf{(2) Camera}: projection, placement, and initial orientation; \textbf{(3) Scene}: layouts, spawns, and task-specific configurations; and \textbf{(4) Control}: input-response gains for camera rotation, movement, driving, and steering where applicable. All four dimensions are jointly randomized during training.
Examples are provided in Figure~\ref{fig:p1-gym-environments}(c).
Appendix~\ref{app:environment-randomization} further specifies the distributions and per-environment settings.

\paragraph{Multi-environment reinforcement learning.}
We jointly optimize the SFT-initialized policy with PPO across randomized environments with different control demands, dynamics, and reward scales.
To limit the effect of reward-scale differences on policy-update magnitudes, we apply \emph{per-environment advantage normalization} to generalized advantage estimates (GAE):
\begin{equation}
\widetilde A_{i,j}=\frac{A_{i,j}-\mu_e}{\sqrt{\sigma_e^2+\epsilon}},\qquad
\mu_e=\frac{\sum_{(i,j)\in\mathcal S_e}A_{i,j}}{N_e},\qquad
\sigma_e^2=\frac{\sum_{(i,j)\in\mathcal S_e}(A_{i,j}-\mu_e)^2}{N_e-1}.
\label{eq:environment-advantage-normalization}
\end{equation}

Here $i$ indexes turns and $j$ response tokens.
For each environment $e$, $\mathcal S_e$ contains its response-token indices in the rollout batch, with $N_e=|\mathcal S_e|$.
We normalize each $A_{i,j}$ for $(i,j)\in\mathcal S_e$ using the corresponding advantage mean $\mu_e$ and sample variance $\sigma_e^2$.
Return-to-go targets accumulate rewards from subsequent turns, allowing later outcomes to inform earlier control decisions.
Appendix~\ref{app:rl-training} and Figure~\ref{fig:browser-rl-infrastructure} detail our rollout infrastructure, reward construction, cross-turn credit, and optimization.

\section{Experiments}

\subsection{Experimental Setup}

\paragraph{Training setup.}
We initialize \playernineb{} and \playerthirtyfiveb{} from their respective Qwen3.5 backbones, each fine-tuned for two epochs on \playerdata{} with action-balanced sampling.
Then, both models are trained with PPO in \playergym{} for 200 steps, with 256 episodes per step distributed equally across the sixteen environments, at most 50 turns per episode. More details of training hyperparameters are provided in 
Appendix~\ref{app:post-training}.

\paragraph{Evaluation.}
Besides in-domain evaluation in \playergym{}, we evaluate out-of-suite transfer on three standardized benchmarks spanning diverse gameplay settings:
\textbf{1).} MineDojo-300~\citep{fan2022minedojo}, with 300 Minecraft tasks and 8 trials per task;
\textbf{2).} ViZDoom-10~\citep{kempka2016vizdoom}, with 10 maps and 50 trials per map; and
\textbf{3).} GameWorld~\citep{ouyang2026gameworld}, with 170 tasks across 34 games and 3 trials per task.
The full protocol comprises 3,410 evaluation episodes per model on environments held out from RL training.
Periodic Minecraft evaluation during RL and the ablations in Section~\ref{sec:environment-analysis} use a fixed 100-task development subset of MineDojo-300, with three trials per task.
Appendix~\ref{app:evaluation-protocols} details benchmark settings.
For baselines, we benchmark against representative game-specific VLAs~\citep{li2025jarvisvla,magne2026nitrogen,yue2026openp2p,guo2026gamewam}, computer-use agents~\citep{bytedance2025uitars15}, and base backbones~\citep{qwen2026qwen35} using their native inference procedures under identical budgets.
Our evaluation compares the base, SFT-initialized, and final \playerone{} models to assess the gains from SFT and subsequent RL on synthetic experience.

\subsection{Main Results}

\paragraph{Open-ended tasks in Minecraft.}
On MineDojo-300, SFT raises success from 9.21\% to 10.54\% for 9B and from 7.79\% to 23.38\% for 35B-A3B.
Interactive learning raises 9B success to 38.33\% (+27.79~pp over SFT), while \playerthirtyfiveb{} reaches 41.04\%, outperforming the top baseline GameScaling-27B (23.29\%) by 17.75~pp.
The number of tasks completed in at least four of eight trials increases from 20 to 136 (and 142 on 35B), demonstrating gains in both task coverage and repeatability.

\begingroup
\setlength{\intextsep}{6pt plus 2pt minus 1pt}
\begin{table}[H]
\setlength{\parskip}{0pt}

\begin{minipage}[t]{\linewidth}
\captionsetup{skip=1pt}

\caption{\textbf{Task success, coverage, and mean steps on MineDojo-300.}
Eight trials per task; brackets show 95\% task-bootstrap confidence intervals.
Mean steps is computed over all episodes, using actual completion steps for successes.}
\label{tab:minedojo-results}
\centering
\normalfont\fontsize{8.5}{9}\selectfont
\setlength{\tabcolsep}{1.5pt}
\sisetup{mode=text,detect-weight=true,reset-text-family=false,reset-text-series=false,reset-text-shape=false,table-number-alignment=center}
\newsavebox{\taskcoverageheading}
\sbox{\taskcoverageheading}{\textbf{\boldmath Task coverage (completed out of 300) $\uparrow$}}
\newlength{\taskcoveragecolwidth}
\setlength{\taskcoveragecolwidth}{\dimexpr(\wd\taskcoverageheading-4\tabcolsep)/3\relax}
\newcommand{\minedojorowhead}[1]{\multirow{2}{*}[-\dimexpr(\aboverulesep+\cmidrulewidth+\belowrulesep)/2\relax]{\textbf{#1}}}
\begin{tabular*}{\linewidth}{@{\extracolsep{0pt plus 1.666667fill}}l
  *{3}{S[table-format=2.2]}
  @{\extracolsep{\fill}\hspace{2\tabcolsep}}l
  >{\centering\arraybackslash}p{\taskcoveragecolwidth}
  @{\extracolsep{0pt}\hspace{2\tabcolsep}}
  *{2}{>{\centering\arraybackslash}p{\taskcoveragecolwidth}}
  @{\extracolsep{\fill}\hspace{2\tabcolsep}}
  S[table-format=4.2]@{}}
\toprule
\minedojorowhead{Method} & \multicolumn{4}{c}{\textbf{\boldmath Success rate (\%) $\uparrow$}}
& \multicolumn{3}{c}{\usebox{\taskcoverageheading}}
& \multicolumn{1}{@{}c}{\minedojorowhead{\boldmath Mean steps $\downarrow$}} \\
\cmidrule(lr){2-5}\cmidrule(lr){6-8}
& {\textbf{Combat}} & {\textbf{Harvest}} & {\textbf{Overall}} & \textbf{\textcolor{black!70}{[95\% CI]}} & {\textbf{\boldmath $\geq$1 of 8}} & {\textbf{\boldmath $\geq$2 of 8}} & {\textbf{\boldmath $\geq$4 of 8}} & \\
\midrule
JARVIS-VLA~\citep{li2025jarvisvla} & 7.56 & 5.00 & 6.71 & \textcolor{black!70}{[5.38, 8.08]} & 95 & 41 & 5 & 950.73 \\
UI-TARS-1.5-7B~\citep{bytedance2025uitars15} & 0.56 & 0.00 & 0.38 & \textcolor{black!70}{[0.13, 0.67]} & 8 & 1 & 0 & 997.34 \\
NitroGen~\citep{magne2026nitrogen} & 0.00 & 0.00 & 0.00 & \textcolor{black!70}{[-]} & 0 & 0 & 0 & 1000.00 \\
Open-P2P-1.2B~\citep{yue2026openp2p} & 0.44 & 0.00 & 0.29 & \textcolor{black!70}{[0.08, 0.50]} & 7 & 0 & 0 & 998.20 \\
GameWAM~\citep{guo2026gamewam} & 14.00 & 9.88 & 12.63 & \textcolor{black!70}{[10.79, 14.54]} & 155 & 79 & 16 & 913.47 \\
GameScaling-27B-S1~\citep{zhang-scaling-games} & 28.94 & 7.75 & 21.88 & \textcolor{black!70}{[19.58, 24.21]} & 182 & 145 & 61 & 844.46 \\
GameScaling-27B-S3~\citep{zhang-scaling-games} & 28.00 & \bfseries 13.88 & 23.29 & \textcolor{black!70}{[20.67, 26.00]} & 185 & 144 & 68 & 833.28 \\
\midrule
Qwen3.5-9B~\citep{qwen2026qwen35} & 11.75 & 4.13 & 9.21 & \textcolor{black!70}{[7.50, 11.00]} & 106 & 59 & 13 & 947.01 \\
Qwen3.5-35B-A3B~\citep{qwen2026qwen35} & 8.69 & 6.00 & 7.79 & \textcolor{black!70}{[6.25, 9.46]} & 99 & 50 & 8 & 953.34 \\
\playerninebsft{} & 12.50 & 6.63 & 10.54 & \textcolor{black!70}{[8.67, 12.50]} & 120 & 63 & 20 & 934.85 \\
\playerthirtyfivebsft{} & 30.31 & 9.50 & 23.38 & \textcolor{black!70}{[20.63, 26.21]} & 172 & 133 & 76 & 839.34 \\
\textbf{\playernineb{}} & 52.31 & 10.38 & 38.33 & \textcolor{black!70}{[35.13, 41.58]} & 209 & 173 & 136 & 706.35 \\
\textbf{\playerthirtyfiveb{}} & \bfseries 55.56 & 12.00 & \bfseries 41.04 & \textcolor{black!70}{[37.83, 44.29]} & \bfseries 224 & \bfseries 182 & \bfseries 142 & \bfseries 705.67 \\
\bottomrule
\end{tabular*}
\par
\end{minipage}
\par
\smallskip
\captionsetup{justification=raggedright,singlelinecheck=false,skip=1pt}
\noindent
\begin{minipage}[t]{0.63\linewidth}
\caption{\textbf{First-person control on ViZDoom-10.}
Mean scaled return over 50 trials per map.
Reward settings and scaling factors are in Appendix~\ref{app:vizdoom-protocol}.}
\label{tab:vizdoom-results}
\end{minipage}\hfill
\begin{minipage}[t]{0.356\linewidth}
\caption{\textbf{Gameplay performance on GameWorld.}
Three trials per task across 170 tasks.}
\label{tab:gameworld-results}
\end{minipage}\par
\vspace{-2pt}
\noindent
\begin{minipage}[t]{0.63\linewidth}
\centering
\normalfont\fontsize{8.5}{9.5}\selectfont
\setlength{\tabcolsep}{1.5pt}
\sisetup{mode=text,detect-weight=true,reset-text-family=false,reset-text-series=false,reset-text-shape=false,table-number-alignment=center,table-text-alignment=center}
\newcommand{\vizdoomhead}[1]{\multicolumn{1}{c}{\bfseries\begin{tabular}[c]{@{}c@{}}#1\end{tabular}}}
\begin{tabular*}{\linewidth}{@{\extracolsep{\fill}}l*{5}{S[table-format=-1.4]}@{}}
\toprule
\textbf{Method} & \vizdoomhead{Precision\\shooting} & \vizdoomhead{Goal\\navigation} & \vizdoomhead{Mobile\\combat} & \vizdoomhead{Survival} & \vizdoomhead{Overall} \\
\midrule
JARVIS-VLA~\citep{li2025jarvisvla} & -0.6270 & -0.1459 & 0.1340 & 0.1565 & -0.1458 \\
UI-TARS-1.5-7B~\citep{bytedance2025uitars15} & -0.6648 & -0.1064 & 0.0587 & 0.1444 & -0.1743 \\
NitroGen~\citep{magne2026nitrogen} & -0.3877 & -0.0732 & 0.0740 & 0.1471 & -0.0793 \\
Open-P2P-1.2B~\citep{yue2026openp2p} & -0.5385 & -0.0174 & 0.6493 & 0.1525 & 0.0602 \\
GameWAM~\citep{guo2026gamewam} & 0.1175 & -0.1285 & \bfseries 1.1860 & 0.1424 & 0.3938 \\
GameScaling-27B-S1~\citep{zhang-scaling-games} & -0.5827 & 0.0280 & 0.2387 & 0.4468 & -0.0083 \\
GameScaling-27B-S3~\citep{zhang-scaling-games} & -0.4282 & 0.3833 & 0.5180 & 0.6069 & 0.2250 \\
\midrule
Qwen3.5-9B~\citep{qwen2026qwen35} & 0.2257 & 0.1155 & 0.4340 & 0.3822 & 0.2975 \\
Qwen3.5-35B-A3B~\citep{qwen2026qwen35} & 0.3376 & 0.1147 & 0.5627 & 0.4909 & 0.3912 \\
\playerninebsft{} & 0.4403 & 0.1839 & 0.5293 & 0.5423 & 0.4361 \\
\playerthirtyfivebsft{} & 0.7376 & 0.3019 & 0.5380 & 0.6065 & 0.5644 \\
\textbf{\playernineb{}} & \bfseries 1.0772 & \bfseries 0.5974 & 0.3033 & 0.7927 & 0.6922 \\
\textbf{\playerthirtyfiveb{}} & 1.0423 & 0.2628 & 0.5960 & \bfseries 0.8979 & \bfseries 0.7236 \\
\bottomrule
\end{tabular*}
\par
\end{minipage}
\hfill
\begin{minipage}[t]{0.356\linewidth}
\centering
\normalfont\fontsize{8.5}{9.5}\selectfont
\setlength{\tabcolsep}{1.5pt}
\sisetup{mode=text,detect-weight=true,reset-text-family=false,reset-text-series=false,reset-text-shape=false,table-number-alignment=center,table-text-alignment=center}
\begin{tabular*}{\linewidth}{@{\extracolsep{\fill}}l*{2}{S[table-format=2.2]}@{}}
\toprule
\multirow{2}{*}{\textbf{Method}} & {\textbf{Success}} & {\textbf{Progress}} \\
& {\textbf{(\%)}} & {\textbf{(\%)}} \\
\midrule
JARVIS-VLA~\citep{li2025jarvisvla} & 6.47 & 17.61 \\
UI-TARS-1.5-7B~\citep{bytedance2025uitars15} & 5.49 & 19.71 \\
NitroGen~\citep{magne2026nitrogen} & 8.82 & 23.41 \\
Open-P2P-1.2B~\citep{yue2026openp2p} & 2.35 & 10.37 \\
GameWAM~\citep{guo2026gamewam} & 5.88 & 16.91 \\
GameScaling-27B-S1~\citep{zhang-scaling-games} & 3.92 & 13.32 \\
GameScaling-27B-S3~\citep{zhang-scaling-games} & 14.90 & 30.72 \\
\midrule
Qwen3.5-9B~\citep{qwen2026qwen35} & 10.39 & 29.22 \\
Qwen3.5-35B-A3B~\citep{qwen2026qwen35} & 13.14 & 33.49 \\
\playerninebsft{} & 13.14 & 31.43 \\
\playerthirtyfivebsft{} & 15.49 & 34.60 \\
\textbf{\playernineb{}} & 15.88 & 34.45 \\
\textbf{\playerthirtyfiveb{}} & \bfseries 17.25 & \bfseries 38.79 \\
\bottomrule
\end{tabular*}
\par
\end{minipage}

\end{table}
\endgroup

\paragraph{Generalization beyond the RL training suite.}
On ViZDoom-10, \playernineb{} raises scaled return from 0.4361 after SFT to 0.6922, and \playerthirtyfiveb{} reaches 0.7236 (+83.7\% over GameWAM at 0.3938).
On GameWorld, progress rises from 31.43\% to 34.45\% on 9B, with \playerthirtyfiveb{} achieving 38.79\% progress.
Together, these results confirm that synthetic control transfers across first-person shooting and browser games without target-domain reinforcement learning.

\begin{figure}[!htb]
\setlength{\abovecaptionskip}{2pt}
\centering
\includegraphics[trim=8.1bp 0 8.1bp 0,clip,width=\linewidth]{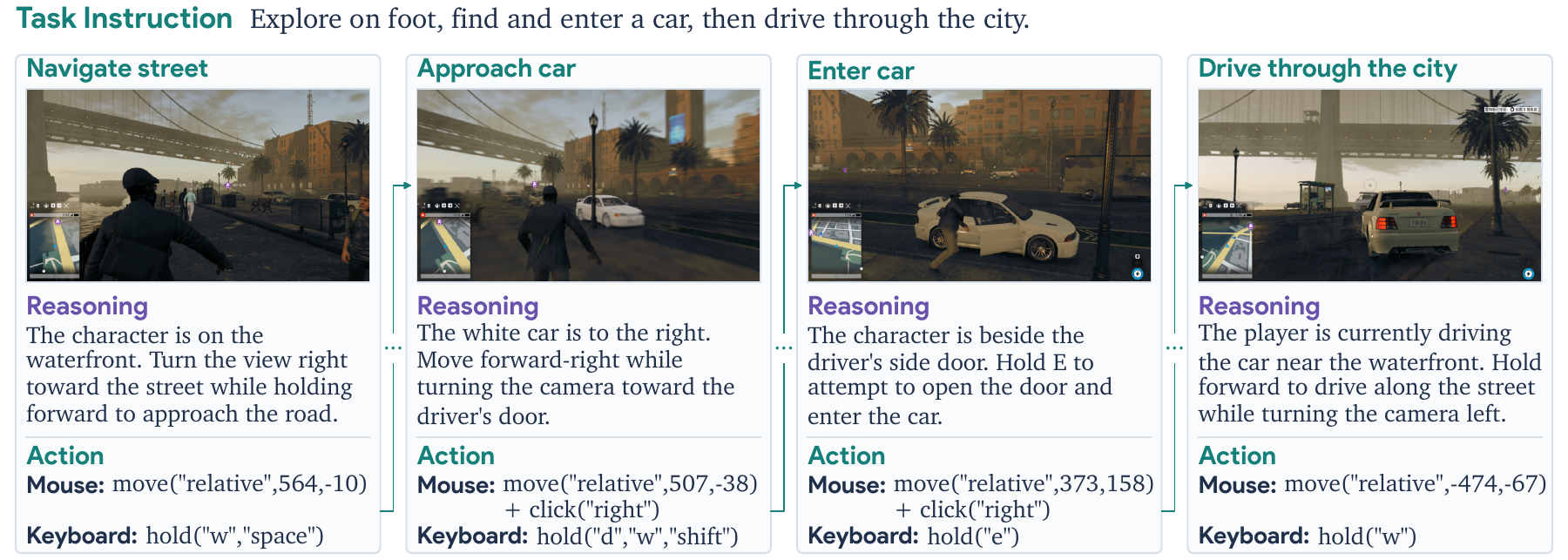}
\caption{Zero-shot generalization of \playernineb{} to the unseen commercial game \textit{Watch Dogs 2}. Conditioned purely on visual feedback, the agent coordinates composite controls across canonical gameplay mechanics: street navigation, target approach, vehicle entry, and driving.}
\label{fig:qualitative-watchdogs2}

\end{figure}

\paragraph{Zero-shot control in an unseen commercial game.}
To evaluate real-world transfer, we deploy \playernineb{} zero-shot in the AAA commercial game \textit{Watch Dogs 2}, which is unseen during both SFT and RL. 
As illustrated in Figure~\ref{fig:qualitative-watchdogs2}, conditioned solely on visual input without privileged game states, the agent orchestrates composite actions across canonical gameplay mechanics: navigating urban streets, approaching an objective, entering a vehicle, and driving.

\section{Analysis}
\label{sec:analysis}

\subsection{Policy Learning and Behavioral Dynamics}
\label{sec:stage-analysis}
\label{sec:learning-dynamics}

\begin{figure}[!htb]

\setlength{\parskip}{0pt}
\setlength{\abovecaptionskip}{2pt}
\centering
\includegraphics[width=\linewidth]{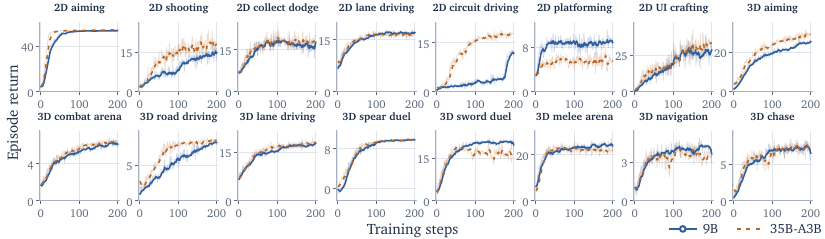}\par
\caption{Per-environment episode returns for 9B and 35B-A3B over 200 RL steps. The curves show differences in learning progress, return scales, and relative model performance across diverse synthetic environments.}
\label{fig:per-environment-learning}

\end{figure}

\noindent
\begin{minipage}[t]{0.48\linewidth}
\vspace{0pt}
\paragraph{Overall training convergence.}
As shown in Figure~\ref{fig:learning-dynamics}, both penalized returns and raw game scores rise steadily across RL steps, with \playerthirtyfiveb{} consistently outperforming \playernineb{}.
Raw game scores track penalized returns, confirming genuine task completion over penalty avoidance.

% The initial run-in heading bypasses the minipage first-paragraph hook.
\makeatletter\@minipagefalse\makeatother
\paragraph{Learning across environments.}
As illustrated in Figure~\ref{fig:per-environment-learning}, learning curves of per-environment returns reflect underlying control complexity: simpler skills such as 2D aiming and shooting converge rapidly, whereas tasks demanding spatial coordination or long-horizon objectives, like 3D navigation and circuit driving, require longer initial exploration.
\end{minipage}\hfill
\begin{minipage}[t]{0.50\linewidth}
\vspace{0pt}
\captionsetup{type=figure}
\centering
\includegraphics[width=\linewidth]{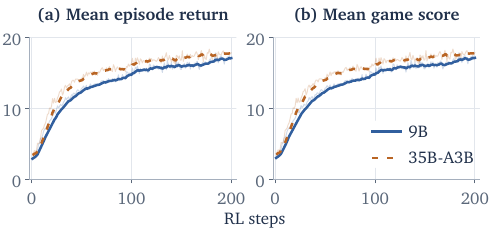}\par
\caption{Training dynamics across 16 environments for 9B and 35B-A3B: (a) mean episode return and (b) mean game score. Faint traces show raw values; bold curves show EMA.}
\label{fig:learning-dynamics}

\end{minipage}\par

\begin{figure}[H]
\setlength{\parskip}{0pt}
\setlength{\abovecaptionskip}{2pt}

\centering
\includegraphics[width=\linewidth]{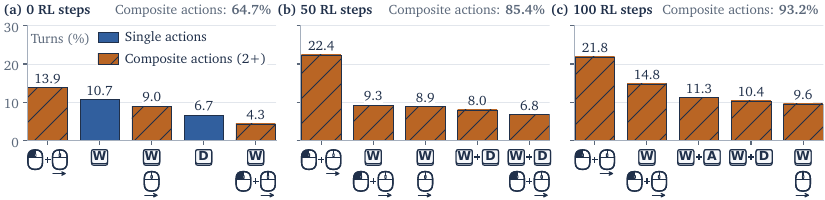}\par
\caption{Evolution of action composition during \playergym{} validation. Bars show each checkpoint's most frequent actions as shares of all turns (\includegraphics[height=2ex]{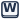} denotes key hold, \includegraphics[height=2ex]{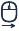} mouse movement, and \includegraphics[height=2ex]{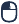} left click). Composite-action use rises from 64.7\% to 93.2\% over 100 RL steps.}
\label{fig:agent-behavior}

\end{figure}

\paragraph{Emergence of composite actions.}
\label{sec:action-composition}
Figure~\ref{fig:agent-behavior} illustrates how the policy's control behavior shifts during RL.
Across \playernineb{}'s in-suite validation trajectories, the proportion of turns utilizing composite actions rises from 64.7\% at step 0 to 93.2\% after 100 RL steps.
Notably, this increase occurs without explicit bonuses for composite inputs, showing that task-directed reinforcement promotes coordinated keyboard--mouse control to satisfy gameplay demands.

\subsection{Initialization and Environment Scaling}
\label{sec:ablation-studies}
\label{sec:environment-analysis}

\noindent
\begin{minipage}[t]{0.48\linewidth}
\vspace{0pt}
We examine the effects of demonstration-based initialization on \playerdata{} and training-suite size in \playergym{}.
All comparisons use a fixed 100-task MineDojo development panel with three trials per task.

\paragraph{Role of supervised initialization.}
As shown in Figure~\ref{fig:env-scaling}(a), training RL directly from the base model ($\text{Base}\rightarrow\text{RL}$) plateaus at only 16.67\%  success, struggling to coordinate controls in the extended action space.
In contrast, initializing from \playerdata{} ($\text{SFT}\rightarrow\text{RL}$) reaches 51.67\%, showing that demonstration priors substantially improve downstream performance under the evaluated RL budget.

\end{minipage}\hfill
\begin{minipage}[t]{0.50\linewidth}
\vspace{0pt}
\captionsetup{type=figure}
\centering
\includegraphics[width=\linewidth]{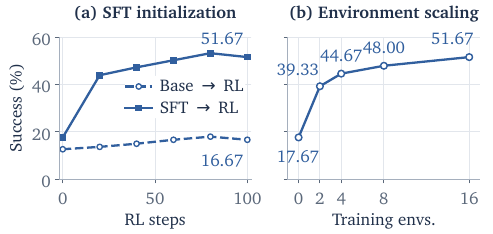}\par
\caption{Initialization and environment scaling for 9B on Minecraft. (a) Effect of SFT initialization. (b) Environment scaling with 25,600 episodes. Both panels use this 100-task development panel with three trials per task.}
\label{fig:env-scaling}

\end{minipage}\par

\paragraph{Scaling the training suite.}
Under a matched budget of 25,600 episodes across 100 RL steps (Figure~\ref{fig:env-scaling}(b)), Minecraft success scales monotonically with suite size, rising from 39.33\% with two environments to 51.67\% with sixteen (vs.\ 17.67\% SFT).
This confirms that cross-game generalization is driven by environment diversity rather than sample volume alone.

\subsection{Retaining General-Purpose Capabilities}
\label{sec:general-retention}

To evaluate whether interactive game post-training preserves base-model capabilities~\citep{liu2026spade}, we audit checkpoints across 13 multimodal settings spanning text (AIME~\citep{aime}, GPQA~\citep{rein2024gpqa}, Reasoning-Gym~\citep{stojanovski2025reasoning}), code (LiveCodeBench~\citep{jain2025livecodebench}), image (MMMU-Pro~\citep{yue2025mmmupro}, MathVision~\citep{wang2024measuring}, BabyVision~\citep{chen2026babyvision}), and video (LVBench~\citep{wang2024lvbench}) under a shared protocol (Table~\ref{tab:general-retention}).
While gameplay-specialized baselines like GameScaling-9B suffer severe capability collapse, dropping by 44--46~pp on text and 28--30~pp on code, \playerone{} broadly retains foundation capabilities: RL trained models remain within 0.99~pp on text and 1.38~pp on image relative to base backbones, and consistently improve code pass@1 by 2.85~pp on 9B and 5.71~pp on 35B-A3B.
Moreover, RL mitigates performance drops observed after SFT, showing that closed-loop control can be acquired while broadly preserving general-purpose capabilities.

\begin{table}[!htb]
\captionsetup{skip=1pt}
\centering
\caption{Retention of base-model general capabilities (\%).
Domain columns report macro-averages across benchmarks and \textbf{Avg} equally weights the four domains.
Bold marks best scores per model group.}
\label{tab:general-retention}
\papertablesize
\setlength{\tabcolsep}{2.5pt}
\begin{tabular*}{\columnwidth}{@{\extracolsep{\fill}}lrrrrr@{}}
\toprule
\textbf{Model} & \textbf{Text} & \textbf{Code} & \textbf{Image} & \textbf{Video} & \textbf{Avg} \\
\midrule
Qwen3.5-9B & \textbf{63.08} & 50.29 & \textbf{48.67} & 32.83 & 48.72 \\
GameScaling-9B-S1 & 16.92 & 22.29 & 26.55 & 21.83 & 21.90 \\
GameScaling-9B-S3 & 18.65 & 20.00 & 26.94 & 20.50 & 21.52 \\
Player-1-9B-SFT & 61.35 & 48.00 & 47.83 & 30.00 & 46.80 \\
Player-1-9B & 62.10 & \textbf{53.14} & 48.41 & \textbf{34.83} & \textbf{49.62} \\
\midrule
Qwen3.5-35B-A3B & \textbf{73.24} & 60.00 & 59.69 & 32.00 & 56.23 \\
Player-1-35B-A3B-SFT & 69.98 & 60.00 & \textbf{60.20} & \textbf{33.17} & 55.84 \\
Player-1-35B-A3B & 73.04 & \textbf{65.71} & 58.32 & 32.83 & \textbf{57.48} \\
\bottomrule
\end{tabular*}

\end{table}

\section{Conclusion}

We presented \playerone{}, a post-training framework for learning transferable gameplay skills from synthetic experience by coupling an extended action space with multi-environment reinforcement learning in \playergym{}.
Evaluations across diverse tasks in Minecraft, ViZDoom, and GameWorld demonstrate that skills acquired in synthetic environments generalize beyond the RL training suite, advancing performance substantially beyond supervised demonstrations.
Further analyses highlight the benefits of supervised initialization and larger training suites, alongside increasingly coordinated controls and broadly preserved general capabilities.

\clearpage
\beginappendix
\section{Gameplay Action Interface}
\label{app:control-interface}

\subsection{Coordinates and Composite Actions}
Table~\ref{tab:control-semantics} specifies the execution semantics of the actions introduced in Section~\ref{sec:action-space}.
For an observation of width $W$ and height $H$, normalized absolute positions and relative displacements map to pixels as
\[
(x,y)\mapsto(xW/1000,yH/1000),\qquad
(\Delta x,\Delta y)\mapsto(\Delta xW/1000,\Delta yH/1000).
\]
The origin is the upper-left corner, and positive displacement points right and down.
One movement uses either absolute or relative coordinates.
A key cannot be both held and tapped in the same step, nor can one mouse button be both held and clicked.

\begin{table}[!htb]

\centering
\caption{Action semantics.
Inputs in a composite action execute concurrently within one step; bindings are restricted to those supported by the environment.}
\label{tab:control-semantics}

\papertablesize
\setlength{\tabcolsep}{3pt}
\begin{tabular}{@{}>{\raggedright\arraybackslash}p{0.16\linewidth}
  >{\raggedright\arraybackslash}p{\dimexpr0.84\linewidth-2\tabcolsep\relax}@{}}
\toprule
\textbf{Component} & \textbf{Semantics} \\
\midrule
Key hold / tap & Hold a set of keys throughout the step, or briefly press and release them. \\
Mouse movement & Move to an absolute screen position or apply a relative displacement. \\
Mouse hold / click & Hold one button throughout the step, or click one button. Different buttons may execute concurrently. \\
Mouse scroll & Apply one upward or downward scroll step. \\
Wait & No operation. Advance one step without keyboard or mouse input. \\
\bottomrule
\end{tabular}
\par
\end{table}

\subsection{Execution and Visual Feedback}
\label{app:execution-feedback}
Relative motion requests input displacement, not a universal camera angle.
For per-pixel camera gain $g$, requested yaw and pitch increments $\delta\psi$ and $\delta\phi$ have magnitudes
\[
|\delta\psi|=gW|\Delta x|/1000,\qquad
|\delta\phi|=gH|\Delta y|/1000,
\]
before native camera limits; mouse displacement is not multiplied by action duration.
The seven mouse-controlled 3D \playergym{} environments use a nominal gain $g_0=0.00196$ radians per observation pixel by default. With Control randomization, the effective gain is $g=g_0L$, where $L$ is defined in Appendix~\ref{app:randomization-sampling}.
Observation geometry and camera constraints remain fixed within a checkpoint comparison. Benchmark-specific settings are given in Appendix~\ref{app:evaluation-protocols}.

\subsection{Observation and Interaction History}
\label{app:interaction-history}
SFT, RL, and evaluation all use two recent visual--action turns and two earlier action-only turns.
In the earlier action-only turns, screenshots are replaced by \texttt{\textless image omitted\textgreater}; reasoning in all retained historical turns is replaced by \texttt{\textless thinking omitted\textgreater}.
The current screenshot and executed historical actions are retained.

\section{\playerdata{}: Construction and Annotation Details}
\label{app:sft-details}
\label{app:player1-data}

\begin{figure}[!t]
\setlength{\parskip}{0pt}
\setlength{\abovecaptionskip}{4pt}

\centering
\includegraphics[width=\linewidth]{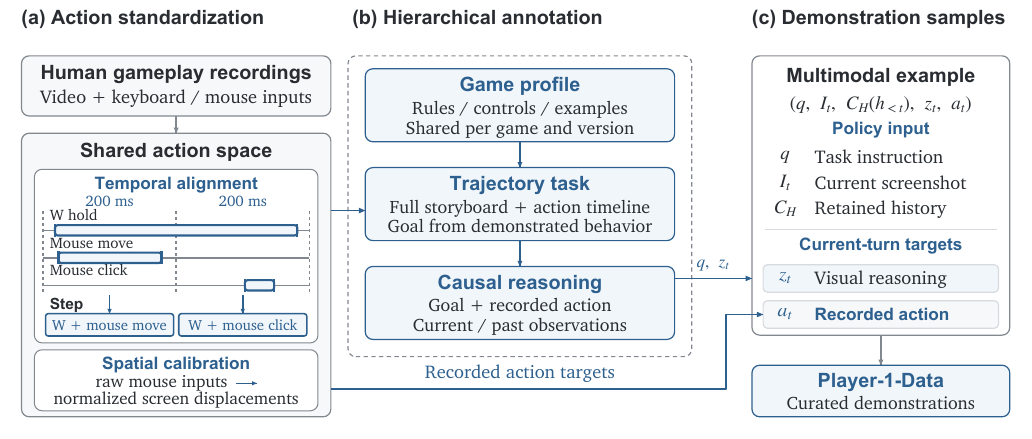}\par
\caption{Overview of the \playerdata{} construction and hierarchical annotation pipeline.
(a)~Action standardization converts raw gameplay inputs into our shared action space via 200~ms temporal alignment and spatial pointer calibration.
(b)~Hierarchical annotation uses Gemini 3.8 Flash with reusable game profiles to generate task goals and causal turn-level visual reasoning $z_t$.
(c)~Demonstration samples pair policy inputs $(q, I_t, C_H(h_{<t}))$ with verified reasoning and recorded human actions $a_t$ for SFT initialization.}
\label{fig:player1-data-construction}
\end{figure}

Figure~\ref{fig:player1-data-construction} illustrates the end-to-end curation and annotation pipeline for \playerdata{}, spanning physical action standardization, Gemini 3.8 Flash hierarchical semantic grounding, and structured demonstration packaging.

\subsection{Sources and Statistics}
Both model sizes are initialized on \playerdata{} (Figure~\ref{fig:p1-data-action-distribution}(a--b)), comprising 6,044 trajectories and 295,097 turns from 2,544 source-qualified recording sessions and 172 games.
We approximate the sampled control hours, which equal the turn count $\times$ 200 ms, or 16.4 hours in total for the full annotated dataset.

\paragraph{Quality selection.}
We automated validity, activity, and repetition checks that complement source-level and manual review of recording completeness, observation--action alignment, and sequence diversity.
The collection contains 5,360 gameplay trajectories with a total of 268,000 turns and 684 in-game UI trajectories with a total of 27,097 turns.

\paragraph{Action statistics.}
Figure~\ref{fig:p1-data-action-distribution}(c) counts the distribution of each action across all samples in the dataset.
Keyboard--mouse composites take precedence. Mouse-only turns are assigned by movement, click, hold, then scroll.
Inputs per step count each key tap/hold and mouse operation separately, with wait as one no-op.
Pointer-mode shares use all turns containing \texttt{mouse\_move}, including keyboard--mouse composites.

\subsection{Action Standardization}
\label{app:temporal-alignment}
As illustrated in Figure~\ref{fig:player1-data-construction}(a), recorded human gameplay inputs are converted into our shared 200-ms action space through temporal interval aggregation and spatial coordinate calibration.

\paragraph{Temporal alignment.}
For continuous gameplay, source timestamps define contiguous intervals of 150--250 ms, projected onto a 200-ms target grid using cumulative boundaries.
Relative input is summed within each source interval.
Keyboard presses shorter than 100 ms become taps, and longer presses become holds on overlapping target steps.
Mouse presses lasting at most 300 ms become clicks.
Each target summarizes compatible recorded inputs for concurrent execution within one control step.

UI trajectories also follow recorded operation boundaries.
Clicks retain their recorded targets, key sequences expand into individual presses, drags use start and endpoint controls with an explicit release, and scroll gestures become directional steps.
Execution uses the shared 200-ms action space.
Sampled control hours use target execution duration, while UI trajectories retain their recorded, operation-dependent timing.

\paragraph{Spatial alignment.}
\label{app:mouse-alignment}
Native relative-mouse input scales vary across data sources and recording setups.
We therefore calibrate these inputs to pixel-equivalent displacements using freely moving GUI pointers from the same recording session.
The mapping also applies to pointer-locked recordings. Camera response is defined in Appendix~\ref{app:execution-feedback}.
For native-input sums $(r_x,r_y)$, coefficients $(k_x,k_y)$, and logical recording size $(W_m,H_m)$, the normalized target is
\[
\Delta x=\operatorname{round}(1000k_xr_x/W_m),\qquad
\Delta y=\operatorname{round}(1000k_yr_y/H_m).
\]

\paragraph{Mouse-move Calibration.}
In-house and D2E calibration uses GUI windows with stable focus and geometry, aligned timestamps, and no pointer clipping or recentering.
For paired native input $r_i$ and pointer displacement $p_i$, window and pooled session estimates are
\[
k_i=\frac{r_i^\top p_i}{r_i^\top r_i},\qquad
\hat k=\frac{\sum_i r_i^\top p_i}{\sum_i r_i^\top r_i}.
\]
D2E uses window-equal estimates for a source-level coefficient; in-house coefficients are recording-specific.
Pixel2Play uses per-axis Huber regression over at least two stable, non-overlapping two-second GUI blocks.
Out-of-range projections are rejected rather than clipped to ensure the action quality. Table~\ref{tab:mouse-mapping} lists the source mappings.
\begin{table}[tbp]

\centering
\caption{Source-specific mouse-action mapping.
Relative inputs use calibration coefficients bound to recording geometry; UI operations use absolute endpoints.}
\label{tab:mouse-mapping}

\papertablesize
\setlength{\tabcolsep}{4pt}
\renewcommand{\arraystretch}{1.08}
\begin{tabular}{@{}>{\raggedright\arraybackslash}p{0.18\linewidth}>{\raggedright\arraybackslash}p{0.78\linewidth}@{}}
\toprule
\textbf{Source} & \textbf{Mapping} \\
\midrule
In-house & Recording-specific calibration coefficients. \\
D2E & Source-level $k_x=k_y=0.66$ for relative input; absolute endpoints for UI operations. \\
Pixel2Play & Hierarchical selection: session GUI fit $\rightarrow$ metadata profile $\rightarrow$ display scale. \\
VPT & $k_x=k_y=2$ in logical $1280\times720$ coordinates for relative input; absolute endpoints for UI operations. \\
Gaming 500h & Recorded absolute positions projected using capture geometry. \\
\bottomrule
\end{tabular}
\par
\end{table}

\subsection{Goal and Reasoning Annotation}
As shown in Figure~\ref{fig:player1-data-construction}(b--c), we employ Gemini 3.8 Flash in a three-stage pipeline to annotate semantic intent and causal reasoning on top of standardized trajectories; Table~\ref{tab:annotation-prompts} summarizes the prompt specifications, inputs, and representative annotated outputs for each stage.
First, we build one reusable game profile, including rules, role and control descriptions, supported bindings, and atomic 200~ms examples, per canonical game, which is shared across data sources, and verify coverage of all recorded controls.
Second, trajectory-level goal inference receives the assembled game profile, full action timeline, and a storyboard of at most 12 frames combining uniform samples (up to 8 frames) with salient event frames ranked by clicks/taps/scrolls, action-state transitions, and peak relative mouse motion.
Third, per-turn causal reasoning conditions strictly on current and past observations alongside the fixed goal and immutable recorded action $a_t$, without showing the annotator future frames. For turns with absolute pointer targets (e.g., UI clicks), a translucent pink target marker is overlaid solely on the annotator's copy of the current frame to aid visual grounding, while SFT training frames remain unmarked.
Crucially, standardized human actions serve directly as target labels $a_t$ and are never generated or modified by the VLM; Table~\ref{tab:annotation-prompts} also shows the resulting serialized \texttt{\textless think\textgreater} and \texttt{\textless tool\_call\textgreater} SFT target format.
Both task and reasoning stages use $\text{temperature}=0$ with strict JSON schemas, together with regular-expression filters that reject responses leaking raw coordinates or tool-call/protocol tokens.

\begin{table}[tbp]

\centering
\caption{Three-stage Gemini 3.8 Flash annotation prompts (verbatim excerpts from the production pipeline), inputs, and real annotated outputs from \playerdata{}.
All stages follow Figure~\ref{fig:player1-data-construction}(b--c); recorded human actions $a_t$ bypass VLM generation.}
\label{tab:annotation-prompts}

\papertablesize
\setlength{\tabcolsep}{4pt}
\renewcommand{\arraystretch}{1.12}
\begin{tabular}{@{}>{\raggedright\arraybackslash}p{0.12\linewidth}
  >{\raggedright\arraybackslash}p{0.18\linewidth}
  >{\raggedright\arraybackslash}p{0.38\linewidth}
  >{\raggedright\arraybackslash}p{\dimexpr0.32\linewidth-6\tabcolsep\relax}@{}}
\toprule
\textbf{Stage} & \textbf{Inputs} & \textbf{Pipeline prompt} & \textbf{Annotated dataset example} \\
\midrule
1. Game profile &
Game name, alternative game names, standardized key/mouse vocabularies, observed recorded controls, one format-reference profile. &
``You create one reusable, source-neutral game and player-role profile [\dots] Describe only stable game rules and keyboard or mouse meanings that are safe to provide before an episode begins. Never mention a particular recording, trajectory, observed or future event, route, score, result, or trajectory-specific goal. [\dots] Return one to three distinct, useful, game-specific in-context examples. [\dots] Return exactly one JSON object matching the supplied schema.'' &
\emph{Cyberpunk 2077:}\newline
\texttt{game\_rules}: ``Follow mission markers through Night City on foot or in vehicles. Manage health, stamina, ammo and cyberware during combat and stealth.''\newline
\texttt{role\_control\_block}: ``WASD: move; shift: sprint [\dots] Relative mouse: look/aim; left click: attack; left hold: automatic/charged fire; right click/hold: aim or scanner zoom [\dots]'' \\
\addlinespace[3pt]
2. Trajectory task &
Game profile, optional source task metadata, full action timeline, $\le 12$-frame storyboard. &
``Infer one task that could be given to a game or interface agent before this demonstration. [\dots] Write a short imperative goal describing what to accomplish, not how to press keys or move the mouse. Never mention coordinates, turn numbers, protocol fields, or unseen future outcomes. The goal must remain coherent with the entire demonstration. [\dots] If the clip is idle, loading, multi-goal, or too ambiguous, return \texttt{review\_required} or \texttt{unavailable}. [\dots] Return only the required JSON object.'' &
\emph{Valorant:}\newline
\texttt{task\_prompt}: ``Deploy Barrier Orb to block the lane and fall back behind cover.'' \\
\addlinespace[3pt]
3. Per-turn causal reasoning &
Game profile, accepted task goal, causal current/past frames (plus target marker for UI clicks), recorded action $a_t$. &
``Produce a causal game or interface annotation with \texttt{visible\_state}, \texttt{immediate\_intent}, \texttt{thinking}, and \texttt{phase}. Semantic text must be useful to a student that sees only the same current-and-past policy frames plus the same fixed game profile and trajectory task. Explain the provided recorded action in that task context; never author or replace an action. [\dots] Do not put coordinates, protocol vocabulary, or future outcomes in semantic text.'' &
\emph{Grand Theft Auto V (in-game UI):}\newline
\texttt{phase}: \texttt{browser\_navigation}.\newline
\texttt{thinking}: ``The objective is to find the listing for 1337 Exceptionalists Way on Dynasty 8 Real Estate. To reach this listing, I need to open the Dynasty 8 Real Estate webpage from the search results. Clicking the highlighted Dynasty 8 Real Estate entry will load the website.''\newline
\emph{Preserved $a_t$:} \texttt{move(abs,386,713)} $+$ \texttt{click(left)}. \\
\midrule
\multicolumn{4}{@{}>{\raggedright\arraybackslash}p{\linewidth}@{}}{%
\textbf{Serialized SFT assistant-target format} (composite keyboard--mouse turn from \emph{Elden Ring}, Fig.~\ref{fig:p1-data-action-distribution}(d); line breaks compacted):\newline
{\fontsize{7.5}{9}\selectfont\texttt{\textless think\textgreater\ The boss's slash animation is completing its swing path above the player. To maintain damage pressure during this recovery window, queueing a heavy attack while advancing ensures the hit connects directly against her hit box.\ \textless/think\textgreater}\newline
\texttt{\textless tool\_call\textgreater\textless function=mouse\_control\textgreater\textless parameter=mouse\_click\textgreater left\textless/parameter\textgreater\textless/function\textgreater\textless/tool\_call\textgreater}\newline
\texttt{\textless tool\_call\textgreater\textless function=key\_control\textgreater\textless parameter=key\_hold\textgreater["w","shift"]\textless/parameter\textgreater\textless/function\textgreater\textless/tool\_call\textgreater}}%
} \\
\bottomrule
\end{tabular}
\par
\end{table}

\section{\playergym{}: Tasks, Rewards, and Randomization}
\label{app:environment-randomization}

\subsection{Tasks and Verifiable Rewards}
\label{app:gym-rewards}
By primary task, the sixteen environments comprise aiming (2), combat (5), driving (4), navigation (4), and UI interaction (1).
Tables~\ref{tab:randomization-2d} and~\ref{tab:randomization-3d} specify their tasks, score components, and API state fields.
Interaction tests, score checks, and human review verify controls, observable transitions, and reward events against human-defined rules.

We designed the shared Three.js browser runtime to provide controllable resets, keyboard--mouse execution, screenshots, and rewards, with fixed-step simulation independent of inference latency.
Each environment defines a cumulative score $S_e(s)$ and supplies the increment $r_{e,t}=S_e(s_{t+1})-S_e(s_t)$ as reward.
Reward computation and evaluation use privileged state, while the policy receives only screenshots and task/control text.
Episodes only end on native completion or failure, or at the turn cap when the task is not completed under the limits.

\begin{table}[tbp]

\centering
\caption{2D environments, controls, and verifiable score components.
State fields are read from \texttt{gameGymAPI.getState().metrics}; \texttt{primary\_score} gives the resulting cumulative score.}
\label{tab:randomization-2d}

\papertablesize
\setlength{\tabcolsep}{3pt}
\renewcommand{\arraystretch}{1.08}
\begin{tabular}{@{}>{\raggedright\arraybackslash}p{0.16\linewidth}>{\raggedright\arraybackslash}p{0.20\linewidth}>{\raggedright\arraybackslash}p{0.24\linewidth}>{\raggedright\arraybackslash}p{\dimexpr0.40\linewidth-6\tabcolsep\relax}@{}}
\toprule
\textbf{Environment} & \textbf{Task and control} & \textbf{Score components} & \textbf{State fields (API)} \\
\midrule
2D aiming & Track and click moving targets & Hits $+0.1\times$ accumulated precision & \mbox{\texttt{targets\_collected}},\newline \mbox{\texttt{precision\_total}} \\
2D shooting & Move, aim, and shoot & Enemies defeated & \mbox{\texttt{kills}} \\
2D collect dodge & Collect while avoiding hazards & Coins collected & \mbox{\texttt{coins\_collected}} \\
2D lane driving & Time lane changes & Coins collected & \mbox{\texttt{coins\_collected}} \\
2D circuit driving & Steer through a route & Checkpoints reached & \mbox{\texttt{checkpoints\_passed}} \\
2D platforming & Run, jump, and dash & Coins, cleared gaps/spikes, and flags & \mbox{\texttt{coins\_collected}},\newline \mbox{\texttt{gaps\_cleared}},\newline \mbox{\texttt{spikes\_cleared}},\newline \mbox{\texttt{flags\_reached}} \\
2D UI crafting & Consult, craft, and equip & Recipe consultations + material reward $+3\times$ crafts + selections $+2\times$ equips & \mbox{\texttt{recipe\_book\_rewards}},\newline \mbox{\texttt{correct\_material\_reward}},\newline \mbox{\texttt{target\_crafts}},\newline \mbox{\texttt{correct\_equipment\_selections}},\newline \mbox{\texttt{equipment\_equipped}} \\
\bottomrule
\end{tabular}
\par
\end{table}

\begin{table}[tbp]

\centering
\caption{3D environments, controls, and verifiable score components.
State fields are read from \texttt{gameGymAPI.getState().metrics}; \texttt{primary\_score} gives the resulting cumulative score.}
\label{tab:randomization-3d}

\papertablesize
\setlength{\tabcolsep}{4pt}
\renewcommand{\arraystretch}{1.08}
\begin{tabular}{@{}>{\raggedright\arraybackslash}p{0.17\linewidth}>{\raggedright\arraybackslash}p{0.245\linewidth}>{\raggedright\arraybackslash}p{0.24\linewidth}>{\raggedright\arraybackslash}p{0.265\linewidth}@{}}
\toprule
\textbf{Environment} & \textbf{Task and control} & \textbf{Score components} & \textbf{State fields (API)} \\
\midrule
3D aiming & Track and hit targets & Target hits & \texttt{hits} \\
3D combat arena & Coordinate movement, aim, and fire & Enemies defeated & \texttt{kills} \\
3D road driving & Follow bends and checkpoints & Checkpoints reached & \texttt{checkpoints\_passed} \\
3D lane driving & Collect and avoid with lane changes & Coins collected & \texttt{coins\_collected} \\
3D spear duel & Track, charge, release, and dodge & Spear hits $-0.5\times$ damage received & \texttt{spear\_hits}, \texttt{damage\_taken} \\
3D sword duel & Approach, aim, and strike & Attack hits + defeats $-0.5\times$ damage received & \texttt{attack\_hits}, \texttt{kills}, \texttt{damage\_taken} \\
3D melee arena & Coordinate facing, range, and movement & Attack hits + defeats $-0.5\times$ damage received & \texttt{attack\_hits}, \texttt{kills}, \texttt{damage\_taken} \\
3D navigation & Reach successive flags & Goals reached & \texttt{goals\_reached} \\
3D chase & Pursue around obstacles & Targets reached & \texttt{targets\_reached} \\
\bottomrule
\end{tabular}
\par
\end{table}

\subsection{Randomization and Sampling}
\label{app:randomization-sampling}
Figure~\ref{fig:perturbation-matrix} shows individual and joint Appearance, Camera, and Scene variation.
The parameters, distributions, and applicable environments for Appearance, Camera, Scene, and Control are specified in Table~\ref{tab:randomization-ranges}.
Specifically, appearance preserves collision geometry and functional roles; Scene varies layouts, spawns, and subsequent entity generation, potentially changing task difficulty.
Appearance and Camera are sampled at reset and fixed within each episode, with triangular modes at native values.
Background lightness and saturation, ambient intensity, and key-light intensity share one sampled quantile, and linked asset components receive compatible color changes.
Orthographic zoom is relative to the native view, including 2D platforming. Initial first-person point-of-view yaw/pitch is distinct from camera-rig pitch.

\paragraph{Control-response sampling.}
Control varies input response independently of camera projection and scene configuration.
For each supported response parameter $k$, we independently sample
\begin{equation}
c_{e,k}\sim\operatorname{Tri}(a_{e,k},1,b_{e,k}),\qquad
c_{e,k}(t)=c_{e,k}(0)\quad\text{throughout the episode}.
\end{equation}
The control seed is independent of domain randomization. Separate parameter substreams set values at reset, unchanged through respawns.
The narrower chasing, platforming, and jump ranges preserve pursuit-speed and traversal constraints.
Collision rules, gravity, attack timing, control bindings, and reward definitions remain fixed.
Table~\ref{tab:training-configuration} specifies the training mixture.

\begin{table}[tbp]

\centering
\caption{Appearance, camera, scene, and control-response randomization configurations.
Multipliers are relative to each game's native setting.
$U$ denotes uniform sampling and $\operatorname{Tri}$ is parameterized by lower bound, mode, and upper bound.
Rig parameters affect only camera components used by the game.
Scene variation is task-specific and may already be present in native settings.}
\label{tab:randomization-ranges}

\papertablesize
\setlength{\tabcolsep}{4pt}
\renewcommand{\arraystretch}{1.08}
\begin{tabular}{@{}>{\raggedright\arraybackslash}p{0.25\linewidth}>{\raggedright\arraybackslash}p{0.09\linewidth}>{\raggedright\arraybackslash}p{0.41\linewidth}>{\raggedright\arraybackslash}p{\dimexpr0.25\linewidth-6\tabcolsep\relax}@{}}
\toprule
\multicolumn{4}{@{}l}{\textbf{(a) Appearance}} \\
\textbf{Visual attribute} & \textbf{Native} & \textbf{Perturbed distribution / choice} & \textbf{Applies to} \\
\midrule
Entity hue shift & $0^\circ$ & Uniform over $\{0,60,120,180,240,300\}^\circ$ & All games \\
Entity saturation & 1 & $U(0.95,1.25)$ & All games \\
Entity lightness & 1 & $U(0.96,1.15)$ & All games \\
Background hue shift & $0^\circ$ & $\operatorname{Tri}(-10,0,10)^\circ$ & All games \\
Background saturation & 1 & $\operatorname{Tri}(0.55,1,1.05)$ & All games \\
Background lightness & 1 & $\operatorname{Tri}(0.32,1,1.04)$ & All games \\
Color temperature & Neutral & Warm / neutral / cool: $0.2/0.6/0.2$ & All games \\
Ambient intensity & 1 & $\operatorname{Tri}(0.35,1,1.15)$ & 3D games \\
Key-light intensity & 1 & $\operatorname{Tri}(0.45,1,1.20)$ & 3D games \\
Key-light azimuth & $0^\circ$ & $-120^\circ/0^\circ/120^\circ$: $0.2/0.6/0.2$ & 3D games \\
Shadow strength & 1 & $\operatorname{Tri}(0.78,1,1)$ & 3D games \\
Visual assets & Native & Compatible non-native variants & All games \\
\bottomrule
\end{tabular}
\par
\begin{tabular}{@{}>{\raggedright\arraybackslash}p{0.25\linewidth}>{\raggedright\arraybackslash}p{0.09\linewidth}>{\raggedright\arraybackslash}p{0.25\linewidth}>{\raggedright\arraybackslash}p{\dimexpr0.41\linewidth-6\tabcolsep\relax}@{}}
\toprule
\multicolumn{4}{@{}l}{\textbf{(b) Camera}} \\
\textbf{Camera parameter} & \textbf{Native} & \textbf{Perturbed distribution} & \textbf{Applies to} \\
\midrule
Orthographic zoom & 1 & $U(0.80,1.00)$ & 2D games \\
FOV offset & $0^\circ$ & $U(-8,8)^\circ$ & 3D games \\
Rig height & 1 & $U(0.90,1.10)$ & 3D games \\
Follow distance & 1 & $U(0.85,1.15)$ & 3D road/lane driving, melee arena, chase \\
Rig pitch offset & 0 & $U(-0.05,0.05)$ rad & 3D except driving \\
Initial yaw: aiming & 0 & $U(-0.10,0.10)$ rad & 3D aiming, 3D combat arena \\
Initial yaw: combat & 0 & $U(-0.05,0.05)$ rad & 3D duel/melee games \\
Initial yaw: navigation & 0 & $U(-0.20,0.20)$ rad & 3D navigation, 3D chase \\
Initial pitch & 0 & $U(-0.05,0.05)$ rad & All three groups \\
\bottomrule
\end{tabular}
\par
\begin{tabular}{@{}>{\raggedright\arraybackslash}p{0.27\linewidth}>{\raggedright\arraybackslash}p{\dimexpr0.73\linewidth-2\tabcolsep\relax}@{}}
\toprule
\multicolumn{2}{@{}l}{\textbf{(c) Scene}} \\
\textbf{Environment} & \textbf{Task-specific variation} \\
\midrule
2D aiming & Target positions, motion, and spatial layouts \\
2D shooting & Connected walls, enemy regions, and spawns \\
2D collect dodge & Connected layouts, coins, and moving obstacles \\
2D lane driving & Initial lane and object sequences \\
2D circuit driving & Circuit geometry, direction, scale, and mirroring \\
2D platforming & Platforms, gaps, hazards, and route widths \\
2D UI crafting & Recipes, quantities, distractors, and inventory layouts \\
3D aiming & Target directions, distances, and spawns \\
3D combat arena & Connected rooms, cover, and spawns \\
3D road driving & Road curvature, phase, and mirroring \\
3D lane driving & Initial lane and object sequences \\
3D spear duel & Arena geometry and opponent spawns \\
3D sword duel & Platforms, cover, and spawns \\
3D melee arena & Connected courtyards, gates, and spawns \\
3D navigation & Obstacles, starts, and goal positions \\
3D chase & Building geometry, starts, and target routes \\
\bottomrule
\end{tabular}
\par
\begin{tabular}{@{}>{\raggedright\arraybackslash}p{0.25\linewidth}>{\raggedright\arraybackslash}p{0.09\linewidth}>{\raggedright\arraybackslash}p{0.22\linewidth}>{\raggedright\arraybackslash}p{\dimexpr0.44\linewidth-6\tabcolsep\relax}@{}}
\toprule
\multicolumn{4}{@{}l}{\textbf{(d) Control}} \\
\textbf{Response multiplier} & \textbf{Native} & \textbf{Perturbed distribution} & \textbf{Applies to} \\
\midrule
Look gain $L$ & 1 & $\operatorname{Tri}(0.8,1,1.2)$ & 3D except driving \\
Movement $M$ & 1 & $\operatorname{Tri}(0.8,1,1.2)$ & 2D shooting, collect dodge, and lane driving; 3D combat arena, lane driving, spear duel, sword duel, melee arena, and navigation \\
Chasing movement $M$ & 1 & $\operatorname{Tri}(0.9,1,1.1)$ & 3D chase \\
Platform movement $M$ & 1 & $\operatorname{Tri}(0.95,1,1.05)$ & 2D platforming \\
Driving $D$ & 1 & $\operatorname{Tri}(0.8,1,1.2)$ & 4 driving games \\
Steering $T$ & 1 & $\operatorname{Tri}(0.8,1,1.2)$ & 3D road driving \\
Jump impulse $J$ & 1 & $\operatorname{Tri}(0.975,1,1.025)$ & 2D platforming; 3D combat arena \\
\bottomrule
\end{tabular}
\par
\end{table}

\begin{figure}[tbp]

\centering
\begingroup
\color{black}
\setlength{\tabcolsep}{1pt}
\newlength{\perturbshotwidth}
\setlength{\perturbshotwidth}{0.11\linewidth}
\setkeys{Gin}{width=\perturbshotwidth}
\newcommand{\perturbshot}[1]{\raisebox{\dimexpr0.5ex-0.5\height\relax}{#1}}
\fontsize{7}{8}\selectfont
\begin{tabular}{@{}>{\raggedright\arraybackslash}m{0.16\linewidth}*{7}{c}@{}}
\toprule
\textbf{Environment} & \textbf{Original} & \textbf{Appearance} & \textbf{Camera} &
\textbf{Scene} & \textbf{Joint 1} & \textbf{Joint 2} & \textbf{Joint 3} \\
\midrule
2D aiming &
  \perturbshot{\includegraphics{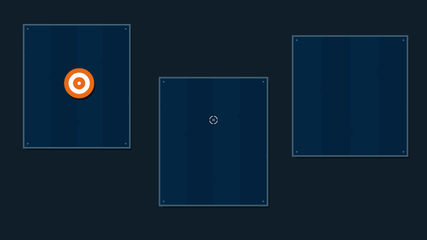}} &
  \perturbshot{\includegraphics{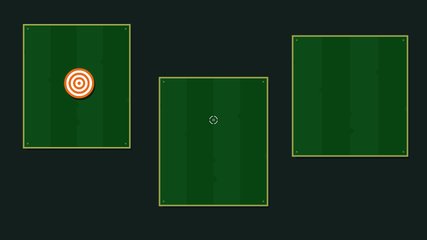}} &
  \perturbshot{\includegraphics{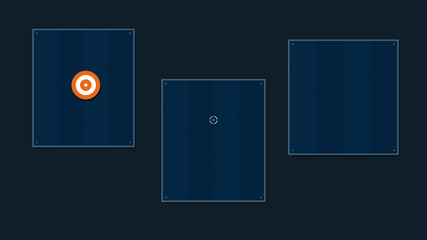}} &
  \perturbshot{\includegraphics{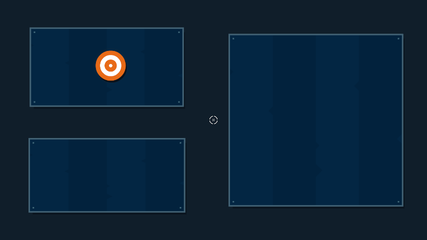}} &
  \perturbshot{\includegraphics{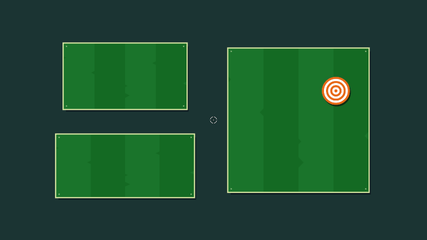}} &
  \perturbshot{\includegraphics{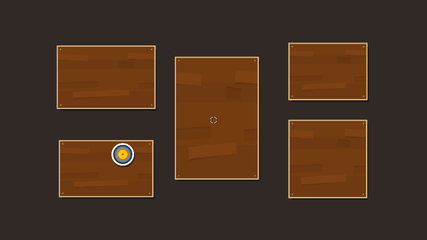}} &
  \perturbshot{\includegraphics{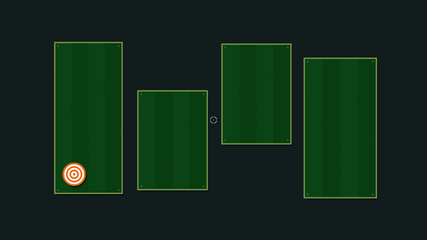}}\\\noalign{\vskip1pt}
2D shooting &
  \perturbshot{\includegraphics{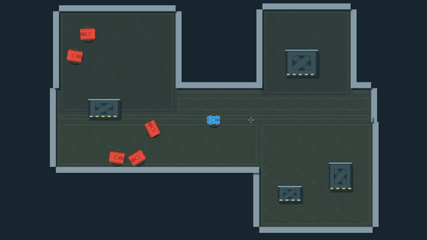}} &
  \perturbshot{\includegraphics{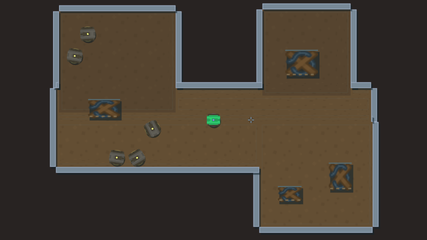}} &
  \perturbshot{\includegraphics{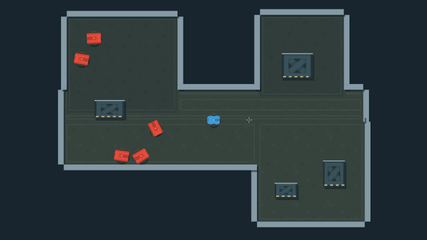}} &
  \perturbshot{\includegraphics{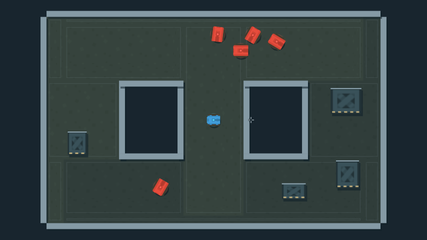}} &
  \perturbshot{\includegraphics{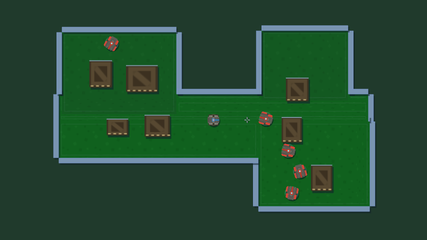}} &
  \perturbshot{\includegraphics{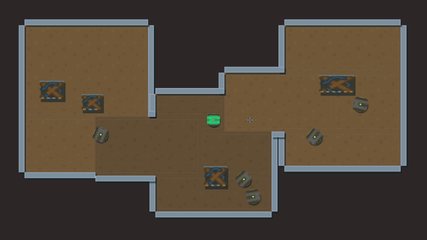}} &
  \perturbshot{\includegraphics{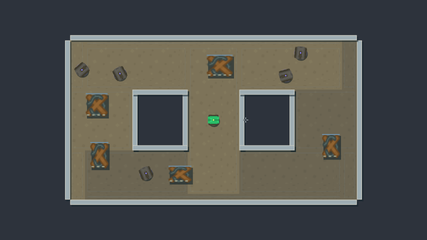}}\\\noalign{\vskip1pt}
2D collect dodge &
  \perturbshot{\includegraphics{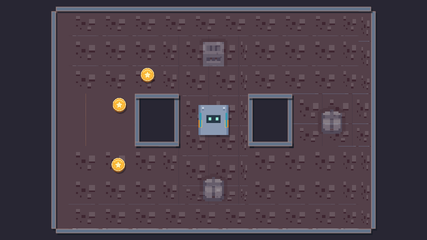}} &
  \perturbshot{\includegraphics{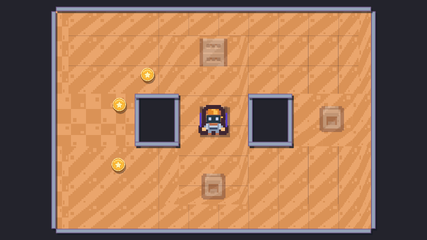}} &
  \perturbshot{\includegraphics{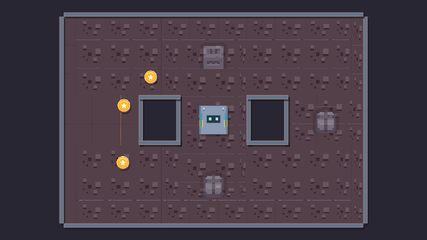}} &
  \perturbshot{\includegraphics{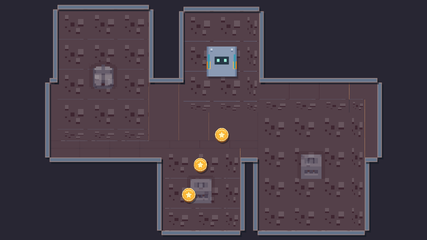}} &
  \perturbshot{\includegraphics{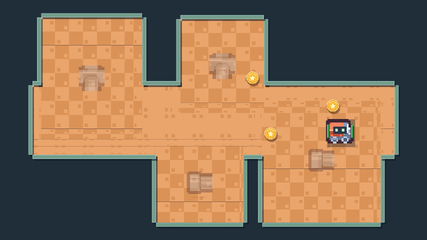}} &
  \perturbshot{\includegraphics{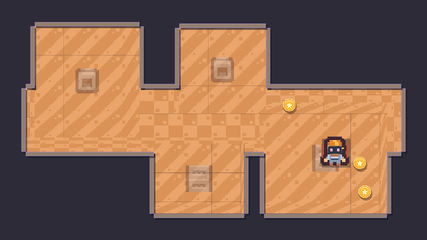}} &
  \perturbshot{\includegraphics{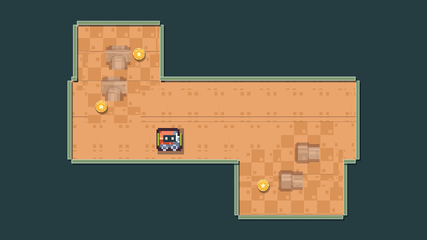}}\\\noalign{\vskip1pt}
2D lane driving &
  \perturbshot{\includegraphics{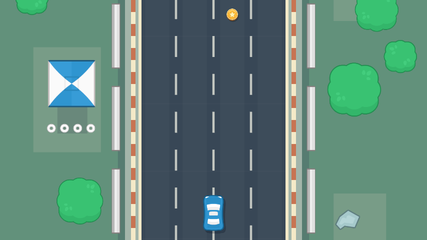}} &
  \perturbshot{\includegraphics{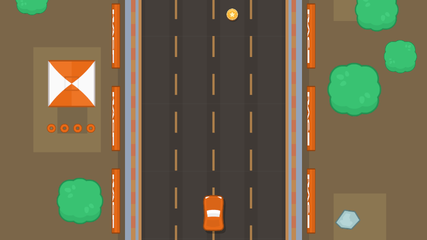}} &
  \perturbshot{\includegraphics{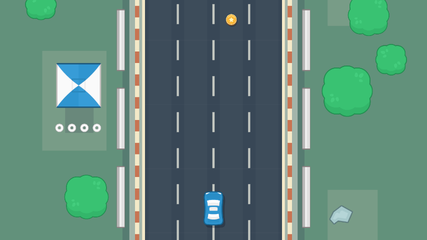}} &
  \perturbshot{\includegraphics{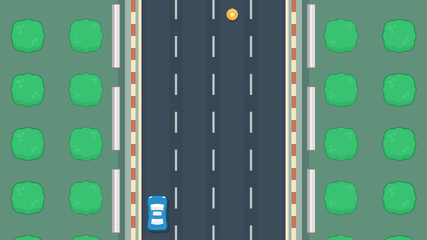}} &
  \perturbshot{\includegraphics{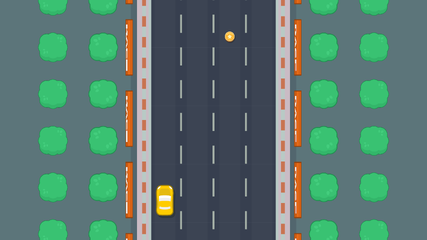}} &
  \perturbshot{\includegraphics{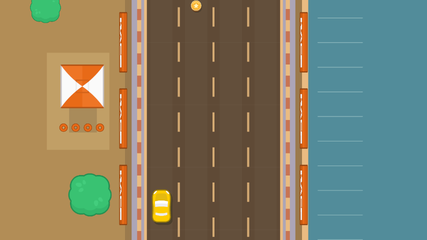}} &
  \perturbshot{\includegraphics{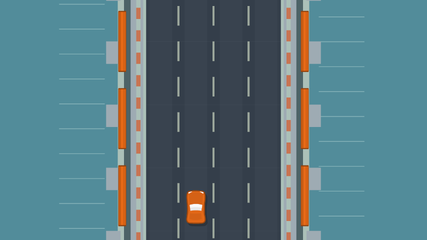}}\\\noalign{\vskip1pt}
2D circuit driving &
  \perturbshot{\includegraphics{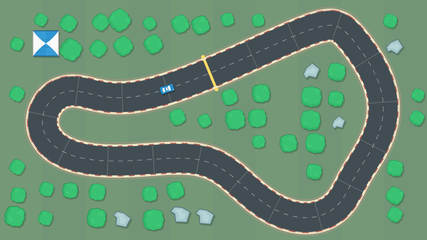}} &
  \perturbshot{\includegraphics{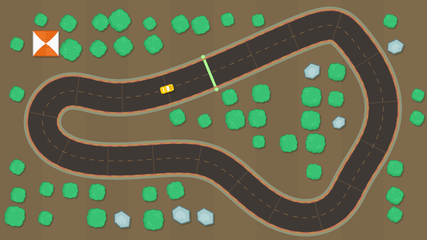}} &
  \perturbshot{\includegraphics{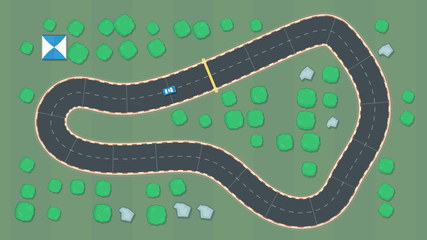}} &
  \perturbshot{\includegraphics{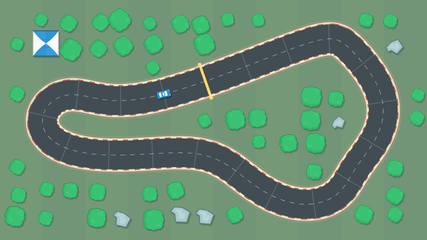}} &
  \perturbshot{\includegraphics{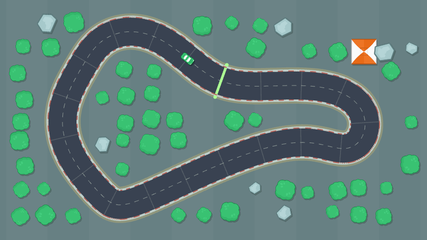}} &
  \perturbshot{\includegraphics{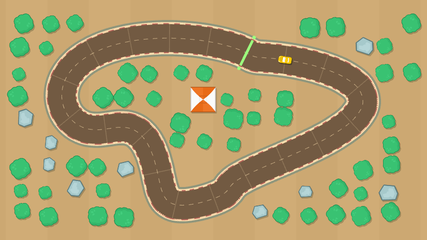}} &
  \perturbshot{\includegraphics{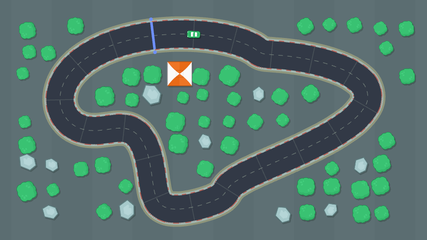}}\\\noalign{\vskip1pt}
2D platforming &
  \perturbshot{\includegraphics{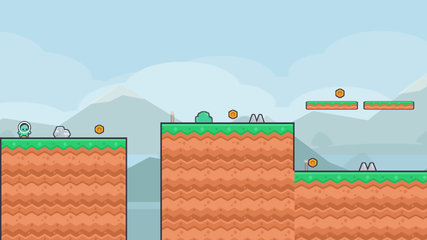}} &
  \perturbshot{\includegraphics{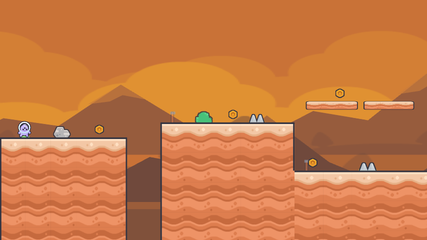}} &
  \perturbshot{\includegraphics{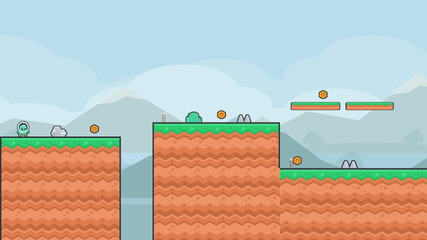}} &
  \perturbshot{\includegraphics{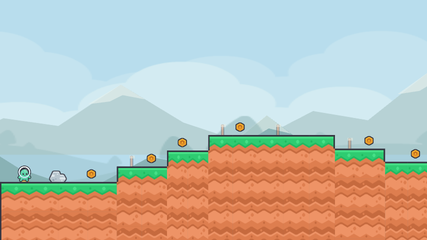}} &
  \perturbshot{\includegraphics{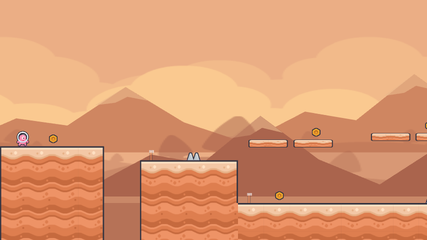}} &
  \perturbshot{\includegraphics{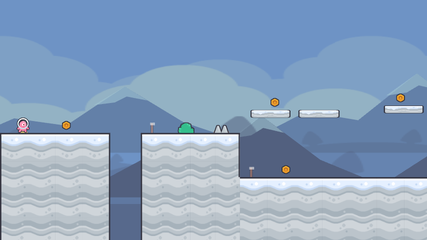}} &
  \perturbshot{\includegraphics{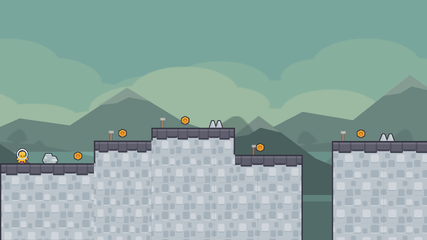}}\\\noalign{\vskip1pt}
2D UI crafting &
  \perturbshot{\includegraphics{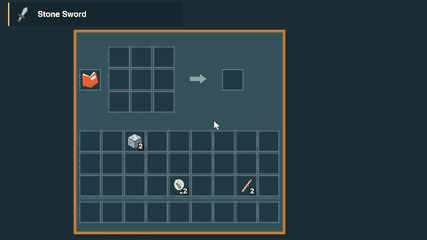}} &
  \perturbshot{\includegraphics{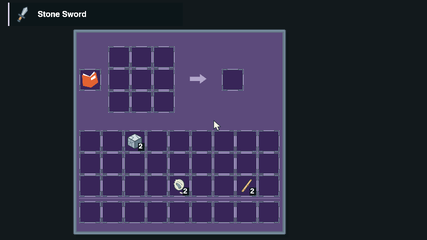}} &
  \perturbshot{\includegraphics{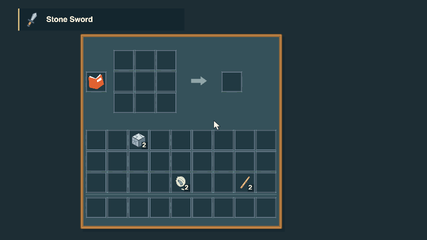}} &
  \perturbshot{\includegraphics{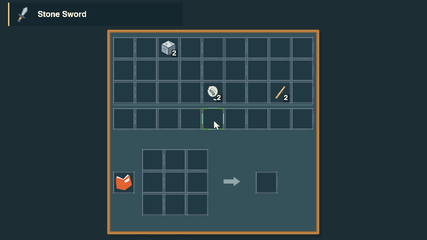}} &
  \perturbshot{\includegraphics{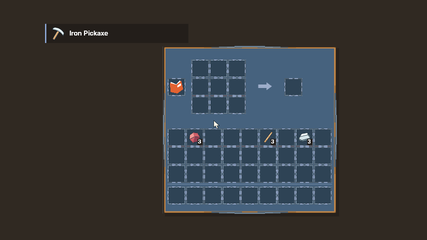}} &
  \perturbshot{\includegraphics{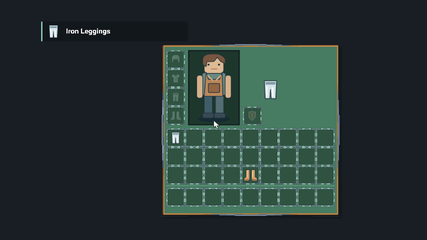}} &
  \perturbshot{\includegraphics{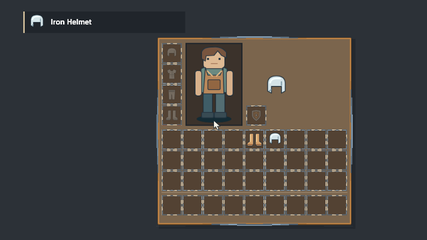}}\\\noalign{\vskip1pt}
3D aiming &
  \perturbshot{\includegraphics{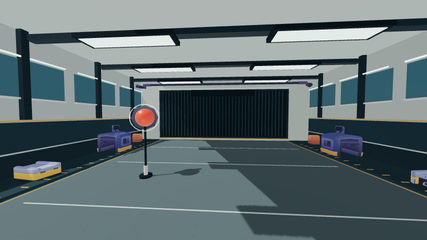}} &
  \perturbshot{\includegraphics{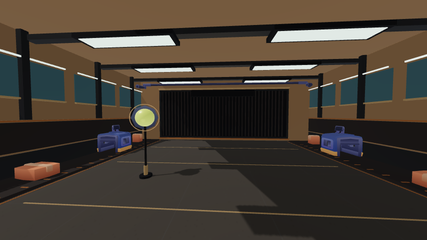}} &
  \perturbshot{\includegraphics{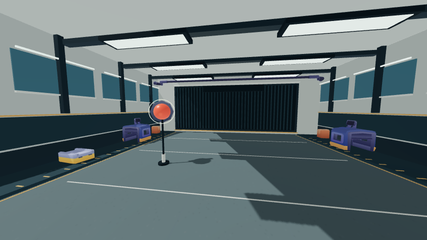}} &
  \perturbshot{\includegraphics{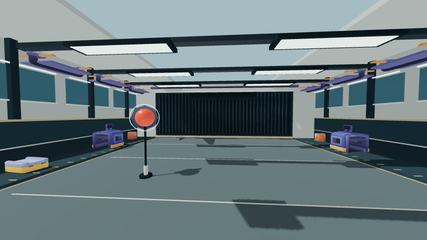}} &
  \perturbshot{\includegraphics{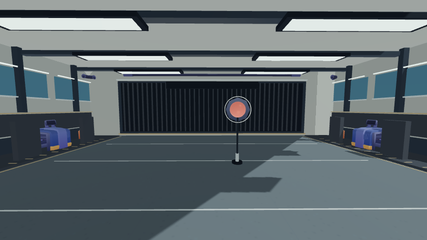}} &
  \perturbshot{\includegraphics{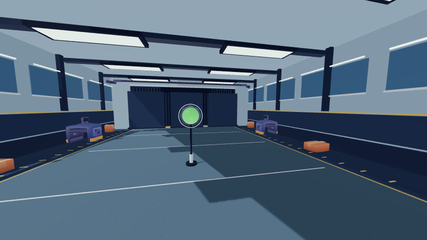}} &
  \perturbshot{\includegraphics{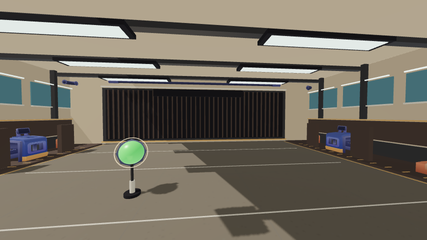}}\\\noalign{\vskip1pt}
3D combat arena &
  \perturbshot{\includegraphics{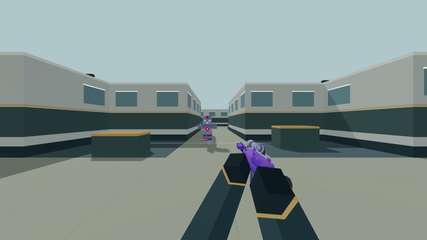}} &
  \perturbshot{\includegraphics{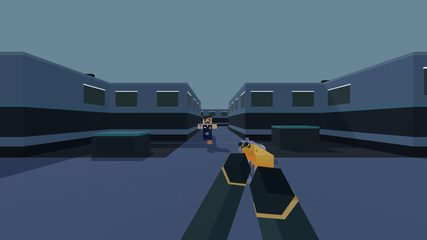}} &
  \perturbshot{\includegraphics{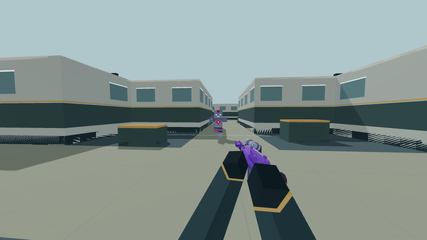}} &
  \perturbshot{\includegraphics{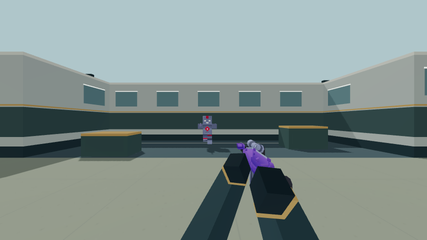}} &
  \perturbshot{\includegraphics{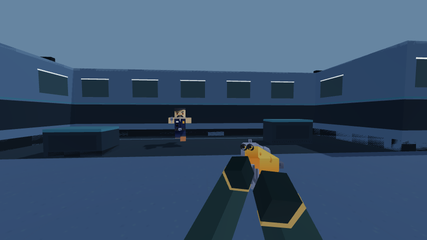}} &
  \perturbshot{\includegraphics{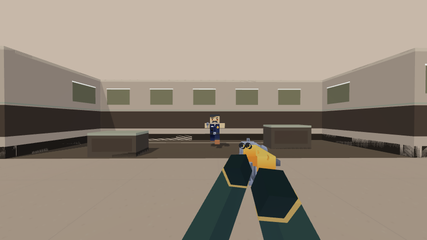}} &
  \perturbshot{\includegraphics{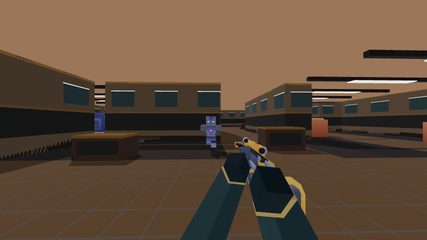}}\\\noalign{\vskip1pt}
3D road driving &
  \perturbshot{\includegraphics{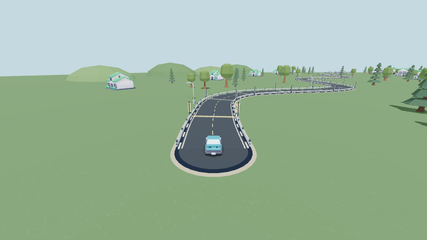}} &
  \perturbshot{\includegraphics{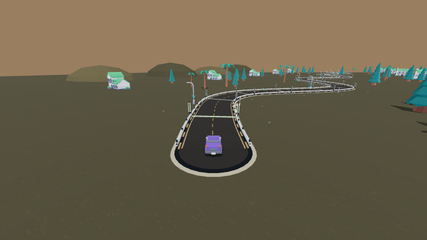}} &
  \perturbshot{\includegraphics{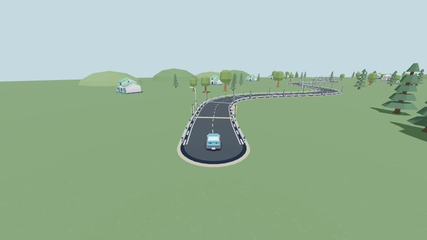}} &
  \perturbshot{\includegraphics{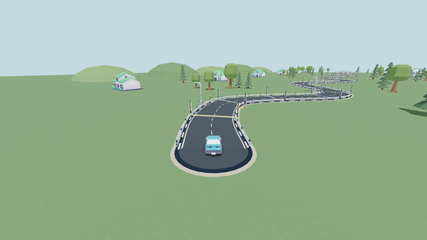}} &
  \perturbshot{\includegraphics{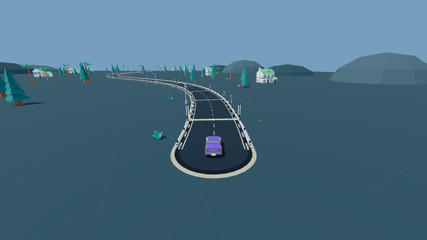}} &
  \perturbshot{\includegraphics{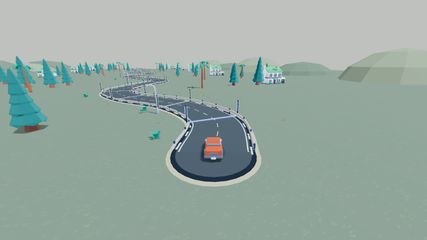}} &
  \perturbshot{\includegraphics{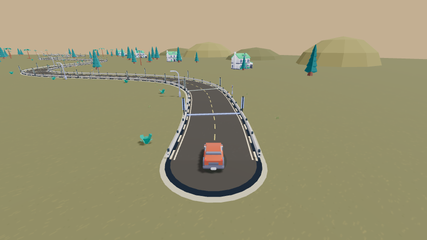}}\\\noalign{\vskip1pt}
3D lane driving &
  \perturbshot{\includegraphics{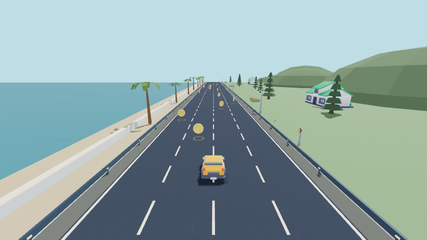}} &
  \perturbshot{\includegraphics{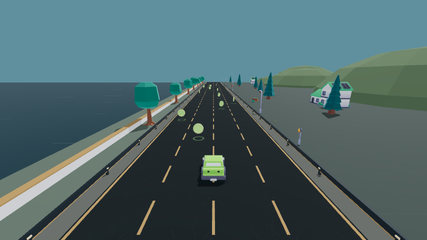}} &
  \perturbshot{\includegraphics{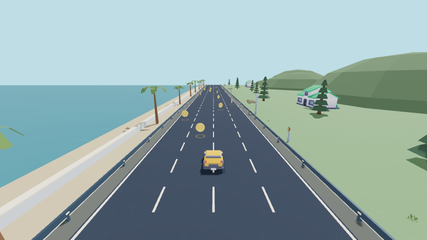}} &
  \perturbshot{\includegraphics{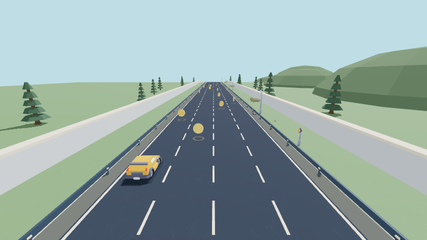}} &
  \perturbshot{\includegraphics{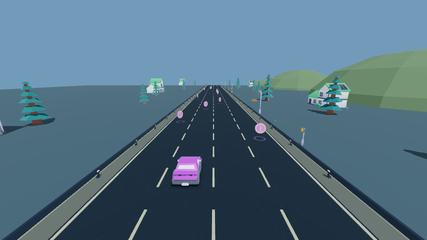}} &
  \perturbshot{\includegraphics{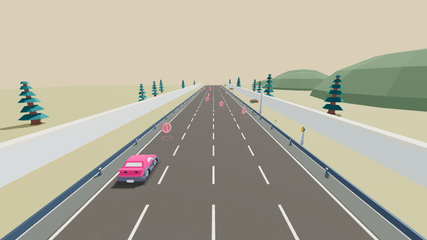}} &
  \perturbshot{\includegraphics{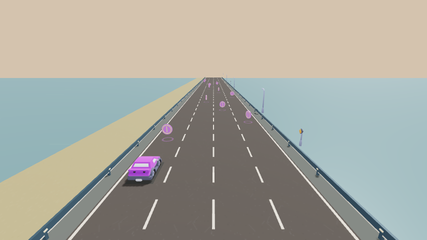}}\\\noalign{\vskip1pt}
3D spear duel &
  \perturbshot{\includegraphics{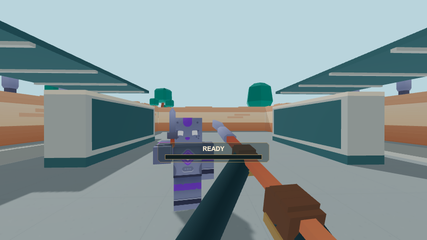}} &
  \perturbshot{\includegraphics{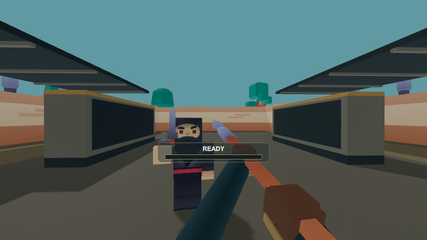}} &
  \perturbshot{\includegraphics{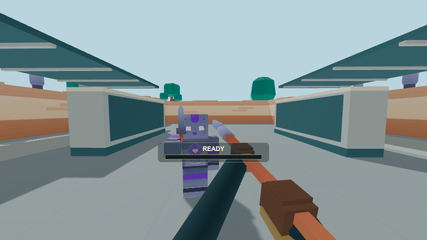}} &
  \perturbshot{\includegraphics{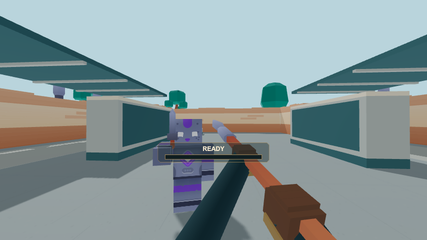}} &
  \perturbshot{\includegraphics{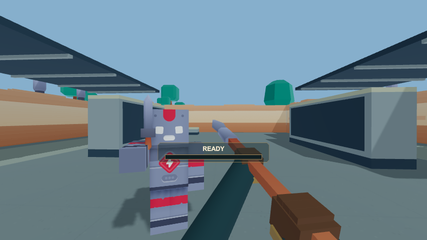}} &
  \perturbshot{\includegraphics{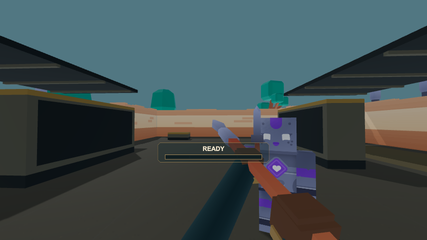}} &
  \perturbshot{\includegraphics{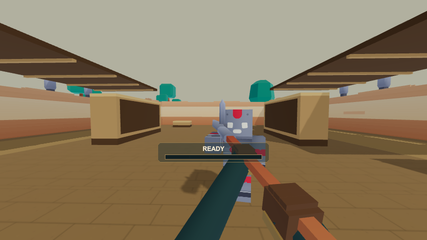}}\\\noalign{\vskip1pt}
3D sword duel &
  \perturbshot{\includegraphics{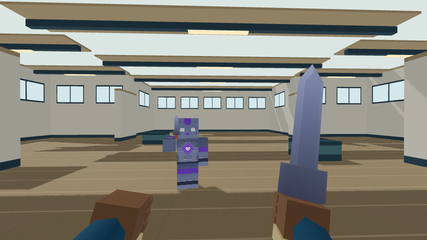}} &
  \perturbshot{\includegraphics{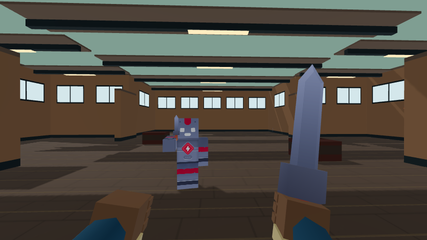}} &
  \perturbshot{\includegraphics{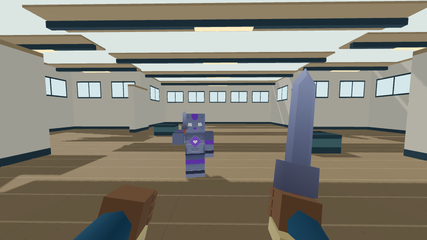}} &
  \perturbshot{\includegraphics{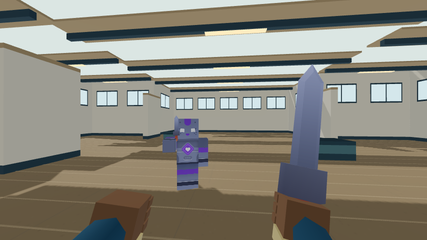}} &
  \perturbshot{\includegraphics{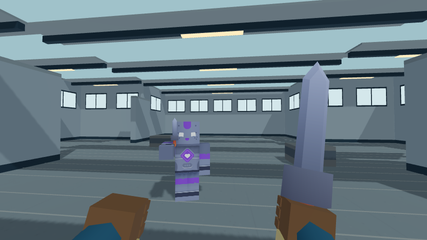}} &
  \perturbshot{\includegraphics{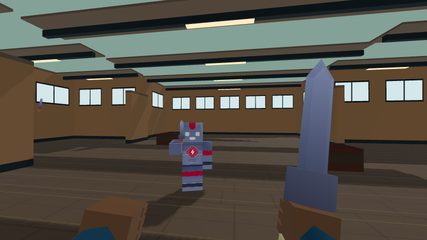}} &
  \perturbshot{\includegraphics{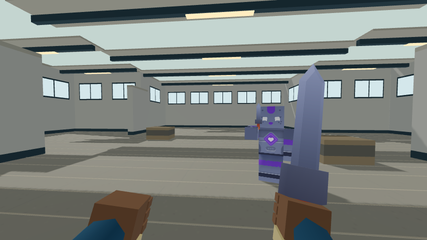}}\\\noalign{\vskip1pt}
3D melee arena &
  \perturbshot{\includegraphics{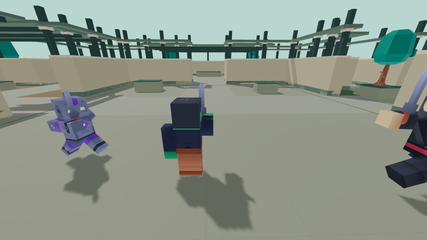}} &
  \perturbshot{\includegraphics{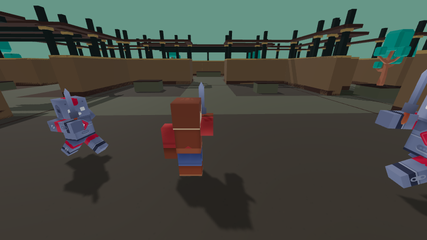}} &
  \perturbshot{\includegraphics{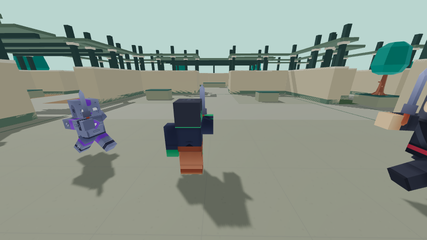}} &
  \perturbshot{\includegraphics{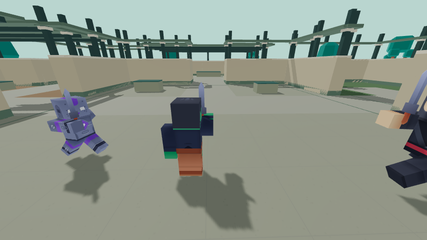}} &
  \perturbshot{\includegraphics{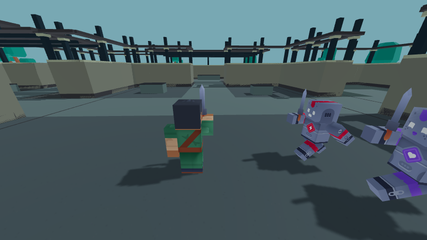}} &
  \perturbshot{\includegraphics{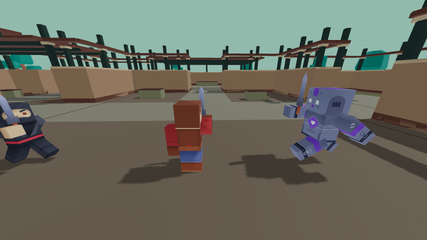}} &
  \perturbshot{\includegraphics{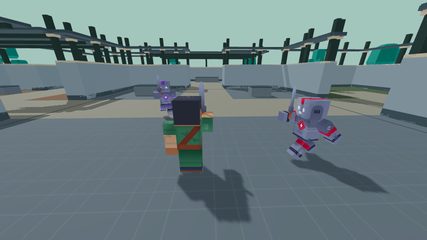}}\\\noalign{\vskip1pt}
3D navigation &
  \perturbshot{\includegraphics{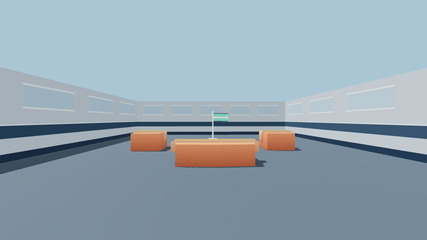}} &
  \perturbshot{\includegraphics{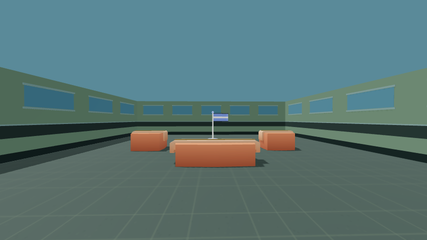}} &
  \perturbshot{\includegraphics{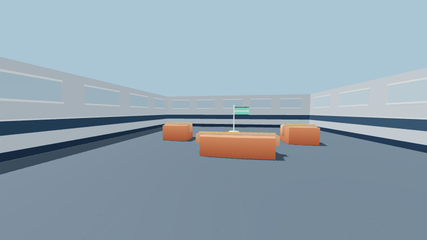}} &
  \perturbshot{\includegraphics{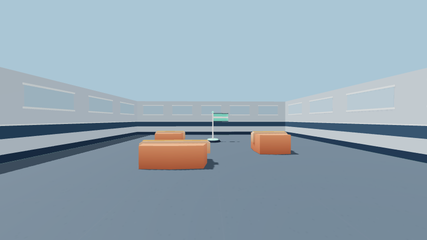}} &
  \perturbshot{\includegraphics{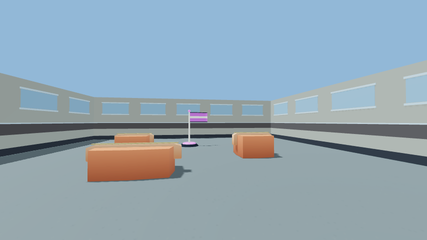}} &
  \perturbshot{\includegraphics{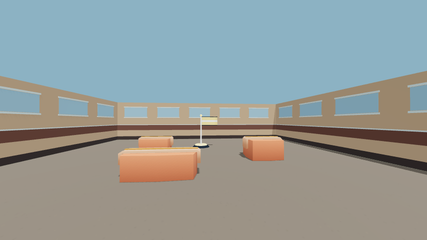}} &
  \perturbshot{\includegraphics{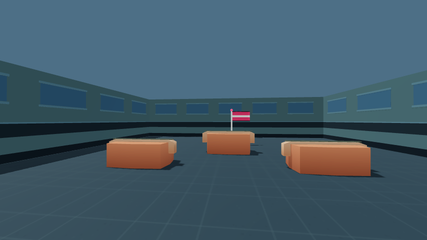}}\\\noalign{\vskip1pt}
3D chase &
  \perturbshot{\includegraphics{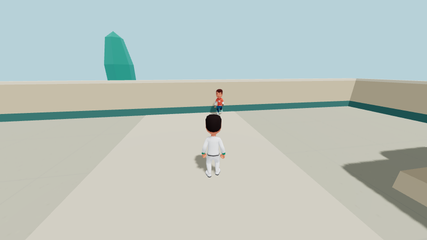}} &
  \perturbshot{\includegraphics{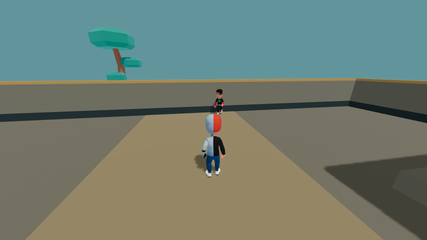}} &
  \perturbshot{\includegraphics{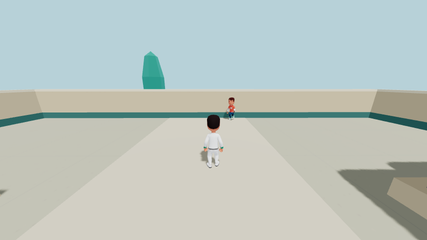}} &
  \perturbshot{\includegraphics{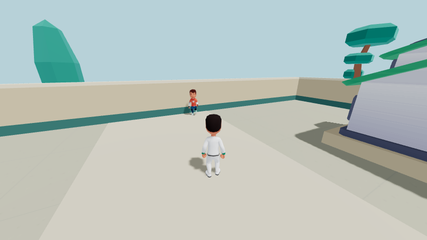}} &
  \perturbshot{\includegraphics{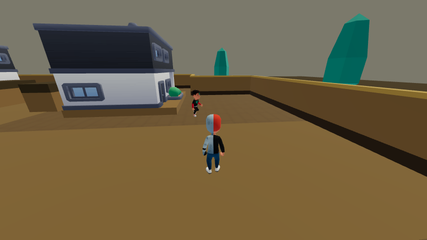}} &
  \perturbshot{\includegraphics{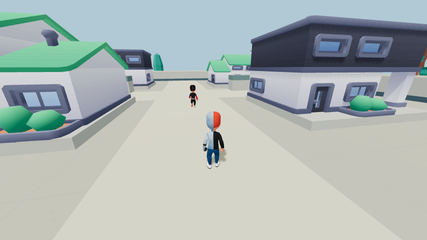}} &
  \perturbshot{\includegraphics{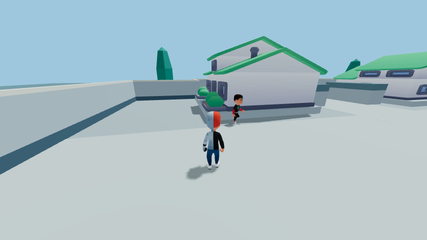}}\\\noalign{\vskip1pt}
\bottomrule
\end{tabular}
\endgroup

\par
\caption{Visual examples of \playergym{} randomization: 16 environments $\times$ seven views.
Original uses default profiles; Appearance, Camera, and Scene change one domain; Joint 1--3 combine these three dimensions.
Control-response variation is omitted because its effects require action sequences rather than static screenshots.}
\label{fig:perturbation-matrix}
\par
\end{figure}

\section{Post-Training Details}
\label{app:post-training}

\subsection{Supervised Fine-Tuning}
\label{app:sft-training}
For supervised fine-tuning, both model sizes use token-level cross-entropy on the current turn's reasoning and action, with the history in Appendix~\ref{app:interaction-history}.

\paragraph{Action-balanced sampling.}
In human gameplay recordings, some actions occupy only a small fraction of playing time but are critical for game progress.
For example, left mouse button inputs account for less than 1\% of valid inputs in the original recordings, which are dominated by keyboard inputs and mouse movement.
In addition to selecting and filtering demonstrations for action coverage during data curation, we use action-balanced sampling to strengthen supervision for these infrequent but important actions.
We classify each turn by its current recorded action into overlapping pools: keyboard, mouse movement, click, hold, any mouse button, keyboard--mouse, movement--click, keyboard--click, and keyboard--movement--click.
Keyboard includes held and tapped keys; keyboard--mouse includes any mouse input, including scrolling.
Other contains the remaining turns, including wait and scroll-only actions.
Each draw selects a pool proportionally to its weight (1 for each control pool; 0.1 for Other), then samples a turn uniformly with replacement.

\subsection{Multi-Environment Reinforcement Learning}
\label{app:rl-training}
During RL, we use synchronous PPO with a dense critic initialized from Qwen3.5-9B for both actor sizes.
Before actor updates begin, the critic undergoes five warmup iterations under the same environment configuration.
Rollouts retain the observations, prompts, responses, masks, and behavior log probabilities used at generation, preserving the same conditioning during optimization.

\paragraph{Reward and cross-turn credit.}
\label{app:cross-turn-credit}
For turn $t$ in environment $e$, we define the penalized step reward and undiscounted return-to-go as
\begin{equation}
\widetilde r_{e,t}=r_{e,t}-p^{\mathrm{invalid}}_{e,t}-p^{\mathrm{length}}_{e,t},
\qquad
G_{e,t}=\sum_{u=t}^{T}\widetilde r_{e,u},
\label{eq:cross-turn-return}
\end{equation}
where $r_{e,t}$ is the game-score increment (Appendix~\ref{app:gym-rewards}) and $T$ the terminal turn.
Training curves average episode returns $G_{e,1}$.
The penalties are
\[
p^{\mathrm{invalid}}_{e,t}=\mathbf{1}[\text{invalid action}],\qquad
p^{\mathrm{length}}_{e,t}=\min\{1,\,0.005\max(0,n_t^{\mathrm{reason}}-64)\},
\]
where $n_t^{\mathrm{reason}}$ counts generated reasoning/prose tokens.
An invalid action executes the runtime's wait fallback.
We apply no additional terminal-failure penalty or task-wise reward rescaling.
With $\gamma=\lambda=1$, return targets are computed over complete trajectories, preserving cross-turn credit.

\paragraph{Advantage normalization and loss reduction.}
We normalize advantages globally in GAE, then within environments using Equation~\eqref{eq:environment-advantage-normalization}, with $N_e\geq2$ and $\epsilon=10^{-8}$; critic return targets are unchanged.
For minibatch $\mathcal B$, let $\mathcal T_i$ be turn $i$'s nonempty response-token set selected by the loss mask.
The clipped PPO loss averages within turns, then across turns:
\begin{equation}
\mathcal L_{\mathrm{policy}}(\theta)
=\frac{1}{|\mathcal B|}\sum_{i\in\mathcal B}
\frac{1}{|\mathcal T_i|}\sum_{j\in\mathcal T_i}
\ell^{\mathrm{PPO}}_{i,j}(\theta;\widetilde A_{i,j}).
\label{eq:turn-mean-policy-loss}
\end{equation}
The critic uses the same reduction. Normalization is token-weighted within environments, whereas losses weight turns equally. Each environment therefore contributes in proportion to its included turns.

\subsection{Training Implementation}
\label{app:training-configuration}
\begin{figure}[!t]
\setlength{\parskip}{0pt}
\setlength{\abovecaptionskip}{4pt}

\centering
\includegraphics[width=\linewidth]{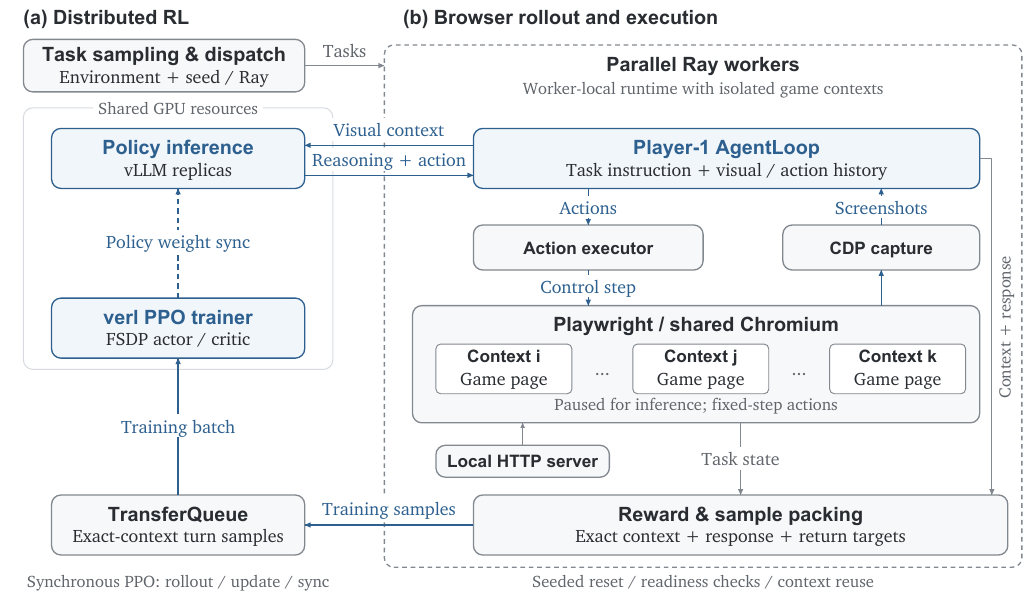}\par
\caption{Overview of the browser-based RL training infrastructure.
(a)~Distributed RL: Ray orchestrates parallel rollout workers streaming exact-context samples via \texttt{TransferQueue} to \textsc{verl}'s FSDP-based PPO trainer, with updated policy weights synchronized back to vLLM inference replicas.
(b)~Browser rollout and execution: \playerone{} agent loops manage isolated Playwright/Chromium contexts, advancing games by fixed control intervals and collecting screenshot observations alongside verifiable state rewards.}
\label{fig:browser-rl-infrastructure}
\end{figure}

Figure~\ref{fig:browser-rl-infrastructure} illustrates our distributed browser-based RL training infrastructure, showing the decoupled coordination between parallel Playwright browser contexts, vLLM policy inference replicas, and synchronous PPO policy optimization.
Table~\ref{tab:training-configuration} summarizes the shared recipe for both model sizes. For each model size, the SFT stage uses 32 NVIDIA H200 GPUs and takes approximately 24 hours, while a 200-step RL run uses 64 H200 GPUs and takes approximately 36 hours.
Both durations include validation.
As depicted in Figure~\ref{fig:browser-rl-infrastructure}(b), browser rollouts use a cluster-wide pool of 32 Ray workers, each managing up to 16 isolated Playwright browser contexts, for a total capacity of 512 concurrent environments.
Each worker maintains a local HTTP server to serve game pages, executes actions via fixed control intervals, and captures screenshots through Chrome DevTools Protocol (CDP).
Games pause during VLM policy inference to prevent wall-clock timing drift.
Upon episode completion or step collection, exact-context turn samples (observations, prompt masks, and behavior log probabilities) are packed alongside return targets and streamed via \texttt{TransferQueue} to \textsc{verl}'s FSDP-based PPO trainer (Figure~\ref{fig:browser-rl-infrastructure}(a)), whose updated weights are periodically broadcast back to vLLM inference replicas.
Browser processes and contexts are reused across episodes to amortize startup costs.

\begin{table}[tbp]

\centering
\caption{Post-training configuration.
Both model sizes share the settings below except for actor architecture.
Each RL step includes rollout collection and PPO updates; critic-only warmup is excluded.}
\label{tab:training-configuration}

\papertablesize
\setlength{\tabcolsep}{4pt}
\renewcommand{\arraystretch}{1.06}
\begin{tabular}{@{}>{\raggedright\arraybackslash}p{0.40\linewidth}
  >{\raggedright\arraybackslash}p{\dimexpr0.30\linewidth-2\tabcolsep\relax}
  >{\raggedright\arraybackslash}p{\dimexpr0.30\linewidth-2\tabcolsep\relax}@{}}
\toprule
\textbf{Setting} & \textbf{\playernineb{}} & \textbf{\playerthirtyfiveb{}} \\
\midrule
Actor / critic & Dense 9B / dense 9B & MoE 35B-A3B / dense 9B \\
\midrule
\multicolumn{3}{@{}l}{\textbf{SFT}} \\
Training trajectories & \multicolumn{2}{l}{6,044} \\
History (visual--action + action-only) & \multicolumn{2}{l}{$2+2$} \\
Sampling & \multicolumn{2}{l}{Pool weights: nine action pools 1 each; Other 0.1} \\
Epoch budget / global batch & \multicolumn{2}{l}{2 / 512 turns} \\
Learning rate / warmup fraction & \multicolumn{2}{l}{$5\times10^{-7}$ / 0.03} \\
\midrule
\multicolumn{3}{@{}l}{\textbf{Interactive learning}} \\
Algorithm / discount / GAE parameter & \multicolumn{2}{l}{PPO / $\gamma=1$ / $\lambda=1$} \\
RL training steps & \multicolumn{2}{l}{200} \\
Critic-only warmup & \multicolumn{2}{l}{5 iterations} \\
History (visual--action + action-only) & \multicolumn{2}{l}{$2+2$} \\
Observation resolution & \multicolumn{2}{l}{480P} \\
Rollout batch / PPO minibatch & \multicolumn{2}{l}{256 episodes / 64 turn responses} \\
Episode cap & \multicolumn{2}{l}{50 turns} \\
Output cap per turn & \multicolumn{2}{l}{256 tokens} \\
Actor learning rate & \multicolumn{2}{l}{$10^{-6}$} \\
Critic learning rate & \multicolumn{2}{l}{$10^{-5}$} \\
Optimizer / learning-rate schedule & \multicolumn{2}{l}{AdamW / constant} \\
AdamW $\beta_1,\beta_2$ / weight decay & \multicolumn{2}{l}{$(0.9,0.999)$ / 0.01} \\
PPO epochs per rollout batch & \multicolumn{2}{l}{1} \\
Gradient-norm clip / PPO clip & \multicolumn{2}{l}{1.0 / 0.2} \\
Reward rescaling & \multicolumn{2}{l}{None} \\
Advantage normalization / loss & \multicolumn{2}{l}{Per environment / token mean, then turn mean} \\
Environment allocation & \multicolumn{2}{l}{Equal across 16 environments} \\
Randomization mixture & \multicolumn{2}{l}{100\% Appearance + Camera + Scene + Control} \\
\bottomrule
\end{tabular}
\par
\end{table}

\subsection{Environment-Scaling Protocol}
\label{app:environment-scaling}
For Section~\ref{sec:environment-analysis}, 9B policies start independently from the same SFT checkpoint, sharing the PPO, history, reward, and randomization settings in Table~\ref{tab:training-configuration}.
Table~\ref{tab:environment-scaling-protocol} gives the suites and matched episode budgets, while interaction lengths and compute may vary within different episodes.
Comparisons share Minecraft development tasks and trial seeds, with SFT as the zero-environment reference.

\begin{table}[tbp]

\centering
\caption{Environment subsets for 9B scaling. All runs use 100 RL steps and 25,600 episodes (256 per step), allocated equally across environments.}
\label{tab:environment-scaling-protocol}

\papertablesize
\setlength{\tabcolsep}{4pt}
\renewcommand{\arraystretch}{1.06}
\begin{tabular}{@{}>{\raggedleft\arraybackslash}p{0.08\linewidth}
  >{\raggedright\arraybackslash}p{0.68\linewidth}
  >{\raggedleft\arraybackslash}p{\dimexpr0.24\linewidth-4\tabcolsep\relax}@{}}
\toprule
\textbf{Envs.} & \textbf{Environment subset} & \textbf{Episodes / env. / step} \\
\midrule
2 & 2D shooting; 3D navigation & 128 \\
4 & 2-env. subset + 2D circuit driving; 3D aiming & 64 \\
8 & 4-env. subset + 2D UI crafting; 2D platforming; 3D sword duel; 3D road driving & 32 \\
16 & Full \playergym{} suite & 16 \\
\bottomrule
\end{tabular}
\par
\end{table}

\section{Evaluation and Results}
\label{app:evaluation-protocols}

\newcommand{\benchmarkframe}[1]{%
  \begingroup
  \setlength{\fboxsep}{0pt}%
  \setlength{\fboxrule}{0.2pt}%
  \setkeys{Gin}{width=\dimexpr0.12\linewidth-2\fboxrule\relax}%
  \fbox{#1}%
  \endgroup
}
\newcommand{\benchmarktrajectory}[3][]{%
  \noindent\begin{minipage}{\linewidth}
  {\small\textbf{#2}}\par
  \if\relax\detokenize{#1}\relax\else
    {\small\normalfont Task instruction: #1\par}%
  \fi
  #3%
  \end{minipage}\par
}

Table~\ref{tab:evaluation-overview} lists the full panels and budgets.
Methods share each panel's tasks, trial seeds, and decision budgets while retaining model-specific preprocessing and inference schedules.
Periodic evaluation during RL steps uses 100 Minecraft tasks with three trials each, ten ViZDoom maps with ten episodes each, and 170 GameWorld tasks with one trial each.
Qwen3.5, \playeronesft{}, and final \playerone{} models are deployed with the history in Appendix~\ref{app:interaction-history}.

\begin{table}[tbp]

\centering
\caption{Evaluation panels, budgets, and reporting units.
Decision limits are per episode (per role in GameWorld).
The three external benchmarks use environments outside \playergym{}.}
\label{tab:evaluation-overview}

\papertablesize
\setlength{\tabcolsep}{3pt}
\begin{tabular}{@{}>{\raggedright\arraybackslash}p{0.17\linewidth}
  >{\raggedright\arraybackslash}p{0.19\linewidth}
  >{\raggedright\arraybackslash}p{0.12\linewidth}
  >{\raggedright\arraybackslash}p{0.19\linewidth}
  >{\raggedright\arraybackslash}p{\dimexpr0.33\linewidth-8\tabcolsep\relax}@{}}
\toprule
\textbf{Panel} & \textbf{Tasks / maps} & \textbf{Trials} & \textbf{Budget} & \textbf{Metrics} \\
\midrule
\playergym{} & 16 games & 128 / game & 50 decisions & Mean episode return \\
MineDojo-300 & 300 tasks & 8 / task & 1,000 decisions & Success; task coverage counts \\
ViZDoom-10 & 10 maps & 50 / map & 150--2,100 decisions & Scaled return \\
GameWorld & 170 tasks; 34 games & 3 / task & 400 decisions & Success; progress \\
\bottomrule
\end{tabular}
\par
\end{table}

\paragraph{Evaluation cost.}
Our full MineDojo-300, ViZDoom-10, and GameWorld protocols allocate up to 3.14 million agent decision turns per model.
Our in-suite validation records average roughly 3,000 input tokens and 100 output tokens per turn, including visual context and generated reasoning.
Using these observed lengths as a workload proxy, a full-budget evaluation would cost approximately USD\$110,000 at GPT-6 Astra's standard API rates of USD\$10 and USD\$50 per million input and output tokens, respectively, before caching discounts.\footnote{\url{https://developers.openai.com/api/docs/models/gpt-6-astra}, accessed Sep 24, 2026.}
Actual costs depend on model-specific tokenization, reasoning, and episode lengths.
Considering the estimated cost, our focus is mainly on representative released gameplay and computer-use policies rather than proprietary API models.

\subsection{Baseline Inference and Action Adapters}
\label{app:baseline-adapters}
We deploy JARVIS-VLA with its action-token decoder and two history entries to predict one action per decision.
UI-TARS-1.5-7B uses three previous screenshots, deterministic decoding, and a 400-token output limit. Its adapter maps pixel-coordinate actions to one benchmark interval, with drag and double-click unsupported.
NitroGen aggregates an 18-action chunk into six benchmark decisions (three native actions per decision), refreshing the image after each chunk.
GameWAM predicts 16 actions and executes eight before replanning, using its Minecraft checkpoint outside Doom and its ViZDoom checkpoint for Doom.
Open-P2P uses visual-only conditioning and session-specific recurrent state.
GameScaling retains its native prompt template and five prior image--action turns, predicting a 200-ms chunk of six key groups and a relative mouse delta mapped to four 50-ms ticks in MineDojo, seven 35-Hz tics with yaw/pitch deltas in ViZDoom, and one 200-ms window in GameWorld.
All temporal inference state is isolated between episodes.

Model inputs retain the released preprocessing for the best practice of each baseline: Minecraft is rendered at $864\times480$. We resize the screenshots to $640\times360$ for JARVIS-VLA and UI-TARS, $256\times256$ for NitroGen, $224\times224$ for GameWAM, and $192\times192$ for Open-P2P, while GameScaling receives native-resolution frames.
JARVIS-VLA and GameWAM retain their native degree-based Minecraft camera actions, while Open-P2P uses an adapter gain of $0.15$ degrees per raw mouse pixel.
Browser-game adapters project native actions onto per-game and per-role bindings.
Benchmark action intervals remain fixed, so a native action chunk can span different simulated durations across benchmarks.

\subsection{\playergym{} In-Suite Validation}
\label{app:in-suite-validation}
In-suite validation uses the seeded panel in Table~\ref{tab:evaluation-overview} with the training randomization settings.
For an episode of $T$ turns in environment $e$, the return is
\[
G_{e,1}=\sum_{t=1}^{T}\left(r_{e,t}-p^{\mathrm{invalid}}_{e,t}-p^{\mathrm{length}}_{e,t}\right).
\]
Here $r_{e,t}=S_e(s_{t+1})-S_e(s_t)$ is the native game-score increment (Appendix~\ref{app:gym-rewards}), and the invalid-action and length penalties are defined in Appendix~\ref{app:rl-training}.
We report mean episode return, averaging trials within each environment and then equally across the sixteen environments.

\begingroup
\setlength{\parskip}{0pt}
\begin{figure}[bhtp]
\centering
\benchmarktrajectory[``hunt a bat in night plains with a iron sword, shield, and a full suite of diamond armors'']{Hunt a Bat in Night Plains with an Iron Sword, Shield, and Diamond Armor}{%
  \benchmarkframe{\includegraphics{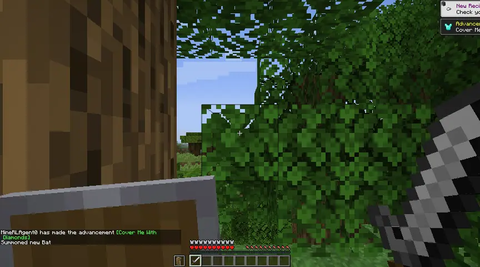}}\hfill%
  \benchmarkframe{\includegraphics{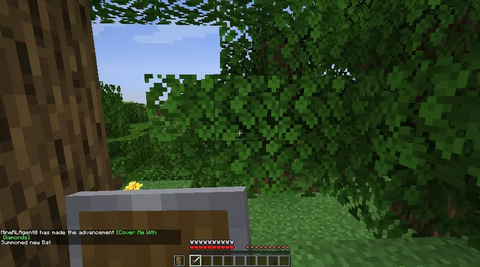}}\hfill%
  \benchmarkframe{\includegraphics{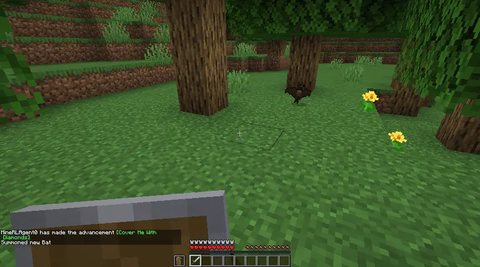}}\hfill%
  \benchmarkframe{\includegraphics{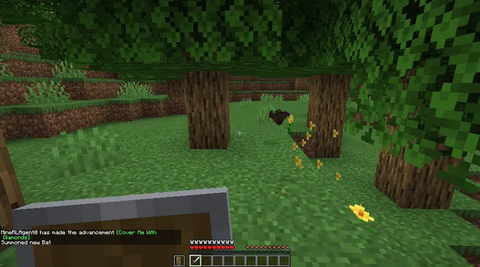}}\hfill%
  \benchmarkframe{\includegraphics{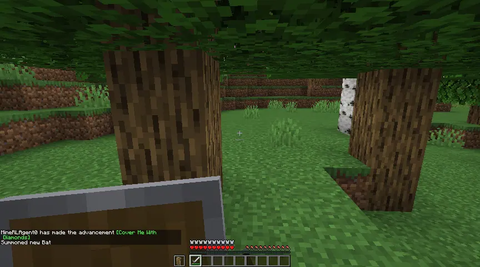}}\hfill%
  \benchmarkframe{\includegraphics{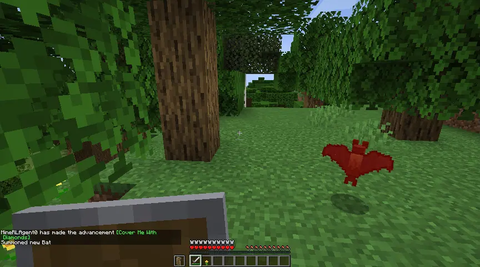}}\hfill%
  \benchmarkframe{\includegraphics{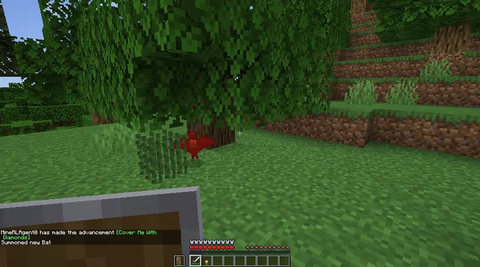}}\hfill%
  \benchmarkframe{\includegraphics{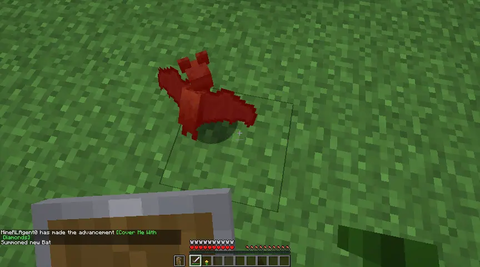}}%
}
\addvspace{4pt}
\benchmarktrajectory[``combat a husk in night desert with a iron sword, shield, and a full suite of diamond armors'']{Fight a Husk in Night Desert with an Iron Sword, Shield, and Diamond Armor}{%
  \benchmarkframe{\includegraphics{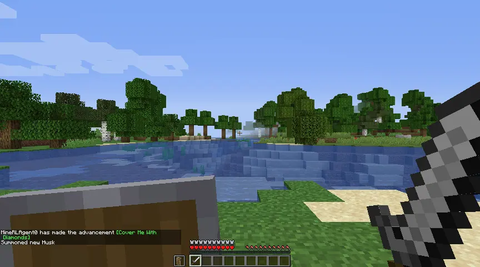}}\hfill%
  \benchmarkframe{\includegraphics{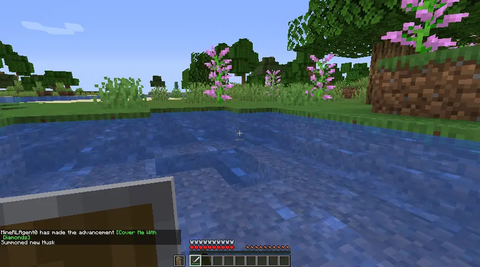}}\hfill%
  \benchmarkframe{\includegraphics{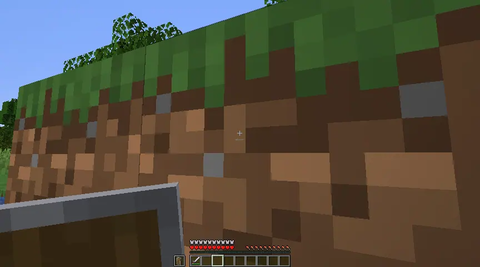}}\hfill%
  \benchmarkframe{\includegraphics{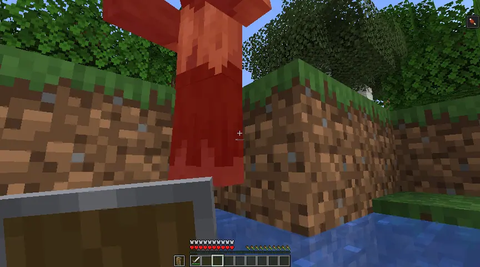}}\hfill%
  \benchmarkframe{\includegraphics{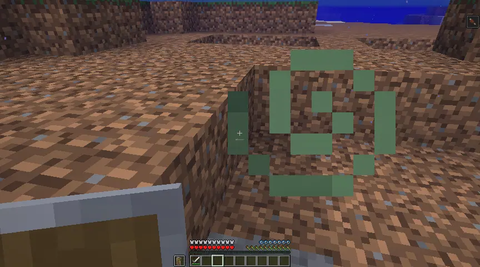}}\hfill%
  \benchmarkframe{\includegraphics{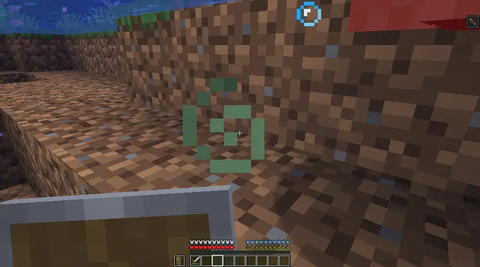}}\hfill%
  \benchmarkframe{\includegraphics{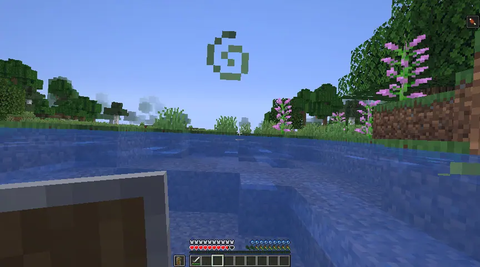}}\hfill%
  \benchmarkframe{\includegraphics{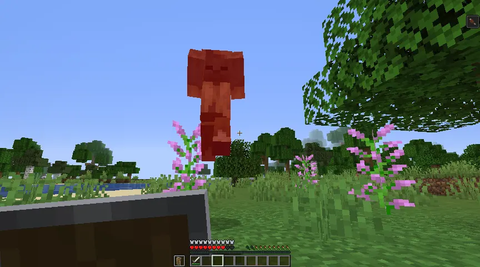}}%
}
\addvspace{4pt}
\benchmarktrajectory[``obtain log in plains'']{Obtain One Log in Plains}{%
  \benchmarkframe{\includegraphics{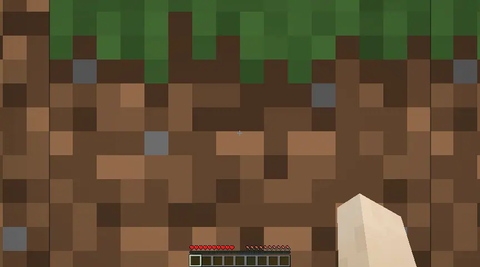}}\hfill%
  \benchmarkframe{\includegraphics{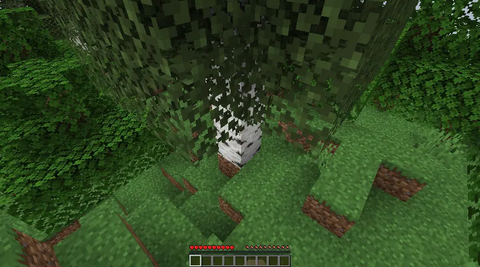}}\hfill%
  \benchmarkframe{\includegraphics{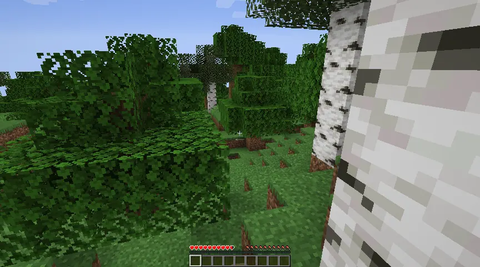}}\hfill%
  \benchmarkframe{\includegraphics{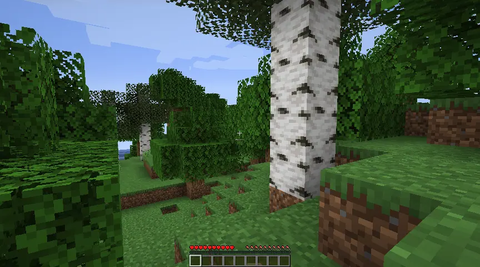}}\hfill%
  \benchmarkframe{\includegraphics{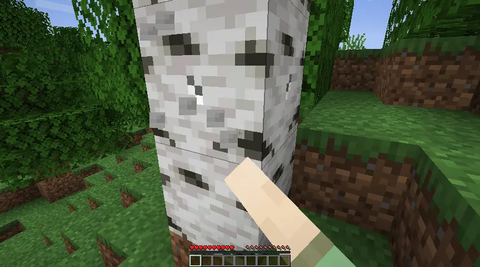}}\hfill%
  \benchmarkframe{\includegraphics{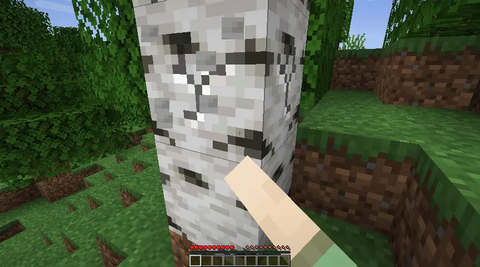}}\hfill%
  \benchmarkframe{\includegraphics{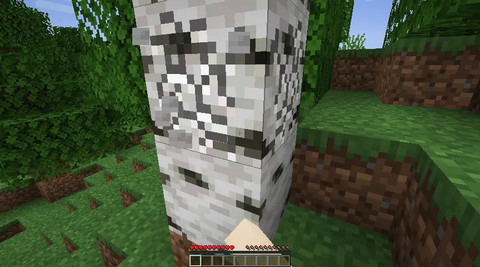}}\hfill%
  \benchmarkframe{\includegraphics{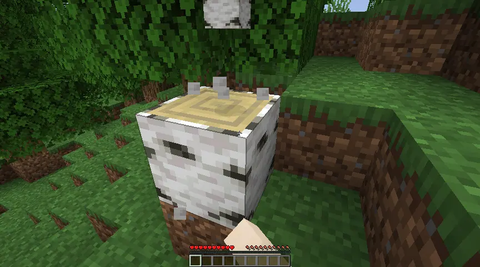}}%
}
\caption{Example tasks and scenes from MineDojo-300.}
\label{fig:minedojo-trajectories}
\end{figure}
\endgroup

\subsection{MineDojo-300}
\label{app:minedojo-protocol}
MineDojo-300 contains 200 Combat and 100 Harvest tasks from MineDojo, ported to MineStudio on Minecraft 1.16.5.
World and task seeds are deterministically derived from each task and trial.
Success is verified after initialization from inventory acquisition for Harvest tasks or target kills for Combat tasks.
We average success over trials within each task, then equally over tasks.
One decision advances a 50-ms simulator tick.
\playerone{}'s camera adapter uses a nominal gain of $0.00196$ radians per observation pixel, subject to native constraints.

We also report \textbf{Task coverage} in $C_m=\sum_{i=1}^{N}\mathbf{1}[s_i\geq m]$ for $m\in\{1,2,4\}$, where $s_i$ counts successes among eight trials for task $i$.
Table~\ref{tab:minedojo-results} reports the number of tasks meeting each threshold, out of 300.
Overall success intervals use 100,000 percentile task-cluster bootstrap replicates, stratified by Combat and Harvest while retaining all eight trials per task and the panel's category weights.
The 95\% intervals quantify within-panel task-sampling uncertainty.
For all-zero outcomes, we report [-].
Mean steps uses the observed completion step for successful episodes and 1,000 for unsuccessful episodes, averaged over all episodes.

\subsection{ViZDoom-10}
\label{app:vizdoom-protocol}

\begin{table}[tbp]

\caption{ViZDoom-10 map categories, raw reward rules, and fixed scales.
Rewards accumulate over an episode; a tic is one simulator tick.
Scaled Return divides the raw episode return by $c_m$.}
\label{tab:vizdoom-map-protocol}

\centering
\papertablesize
\setlength{\tabcolsep}{3pt}
\begin{tabular}{@{}>{\raggedright\arraybackslash}p{0.155\linewidth}
  >{\raggedright\arraybackslash}p{0.22\linewidth}
  >{\raggedright\arraybackslash}p{0.46\linewidth}
  >{\raggedleft\arraybackslash}p{0.12\linewidth}@{}}
\toprule
\textbf{Category} & \textbf{Map} & \textbf{Raw reward rule} & \textbf{Scale $c_m$} \\
\midrule
\multirow[t]{3}{=}{Precision shooting} & Basic & $+106$ for eliminating the target; $-5$ per shot; $-1$ per tic. & 100 \\
 & Defend the Center & $+1$ per kill; $-1$ on death. & 10 \\
 & Defend the Line & $+1$ per kill; $-1$ on death. & 10 \\
\midrule
\multirow[t]{2}{=}{Goal navigation} & Deadly Corridor & Forward position along the corridor; $+1000$ for collecting armor; $-100$ on death. & 1000 \\
 & My Way Home & $+1$ for reaching the goal; $-0.0001$ per tic. & 1 \\
\midrule
\multirow[t]{3}{=}{Mobile combat} & Battle 1 & $+1$ per kill. & 10 \\
 & Battle 2 & $+1$ per kill. & 10 \\
 & Deathmatch & Enemy-dependent kill bonuses, with 1, 3, 3, 4, 3, and 10 for the six enemy types. & 10 \\
\midrule
\multirow[t]{2}{=}{Survival} & Health Gathering & $+1$ per tic alive; $-100$ on death. & 2100 \\
 & Health Gathering Supreme & $+1$ per tic alive; $-100$ on death. & 2100 \\
\bottomrule
\end{tabular}
\par
\end{table}

Table~\ref{tab:vizdoom-map-protocol} lists the ten selected maps, control categories, native rewards, and fixed return scales.
ViZDoom 1.3.0 provides $640\times480$ RGB observations.
Each decision advances two simulator tics. Basic permits 150 decisions, Deathmatch 2,100, and the other maps 1,050, subject to earlier native termination.
\playerone{}'s camera adapter uses $0.00196$ radians per observation pixel with native pitch constraints.

From raw return $R_{m,s}$ and map-specific scale $c_m>0$, we compute map, category ($\mathcal M_c$), and overall scaled returns:
\[
S_m=\frac{1}{N}\sum_{s=1}^{N}\frac{R_{m,s}}{c_m},\qquad
S_c=\frac{1}{|\mathcal M_c|}\sum_{m\in\mathcal M_c}S_m,\qquad
S_{\mathrm{overall}}=\frac{1}{10}\sum_{m=1}^{10}S_m.
\]
Evaluation uses $N=50$ fixed seeds per map, with identical scales across models.
The survival scale of 2,100 corresponds to the native time-reward budget, with death penalties retained.
Observation geometry, timing, and camera constraints are matched within each backbone.

\begingroup
\setlength{\parskip}{0pt}
\begin{figure}[tbp]
\centering
\benchmarktrajectory[``Fight through the corridor and reach the green armor at the far end while staying alive.'']{Deadly Corridor}{%
  \benchmarkframe{\includegraphics{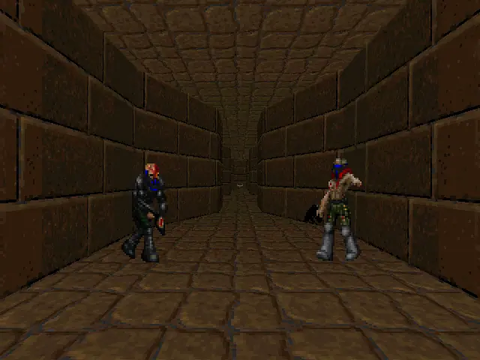}}\hfill%
  \benchmarkframe{\includegraphics{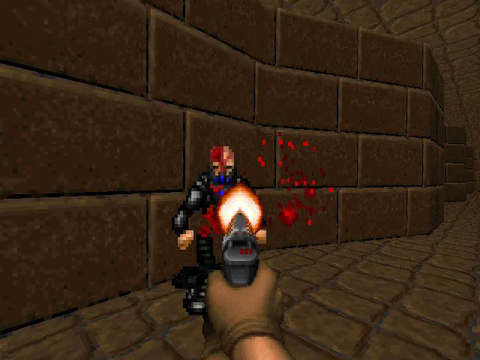}}\hfill%
  \benchmarkframe{\includegraphics{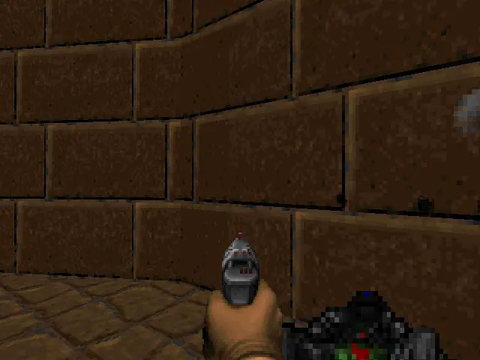}}\hfill%
  \benchmarkframe{\includegraphics{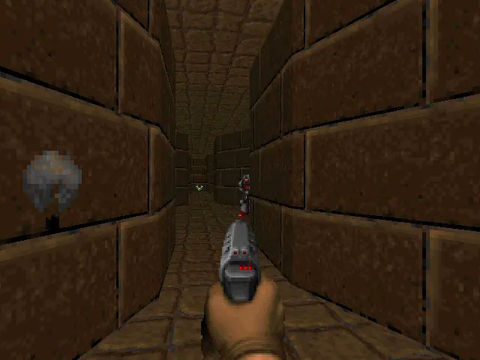}}\hfill%
  \benchmarkframe{\includegraphics{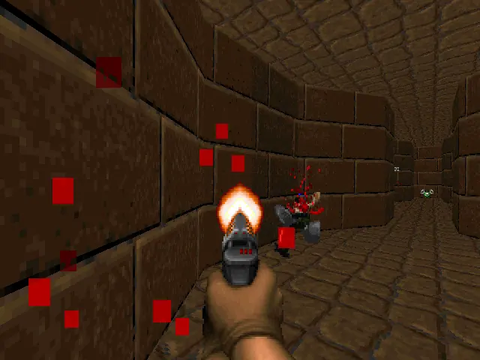}}\hfill%
  \benchmarkframe{\includegraphics{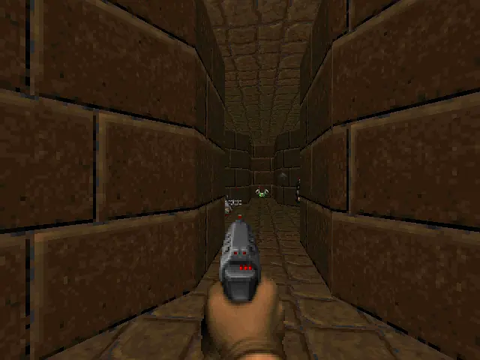}}\hfill%
  \benchmarkframe{\includegraphics{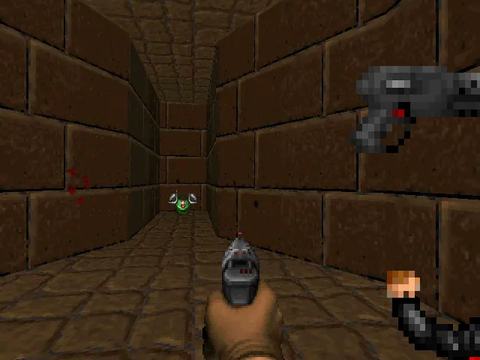}}\hfill%
  \benchmarkframe{\includegraphics{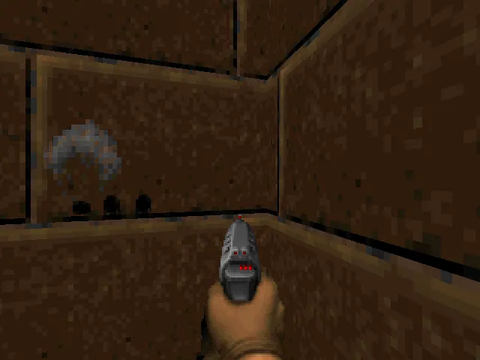}}%
}
\addvspace{4pt}
\benchmarktrajectory[``Fight the monsters and maximize the total episode reward. Move through the arena, keep enemies in view, and fire accurately while staying alive.'']{Battle 2}{%
  \benchmarkframe{\includegraphics{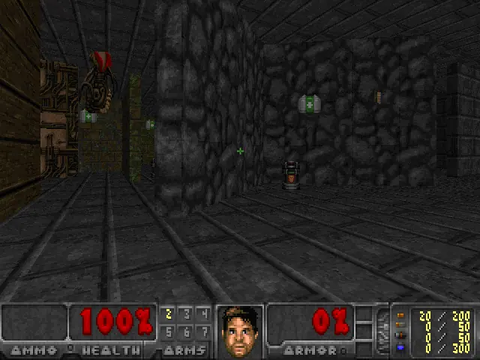}}\hfill%
  \benchmarkframe{\includegraphics{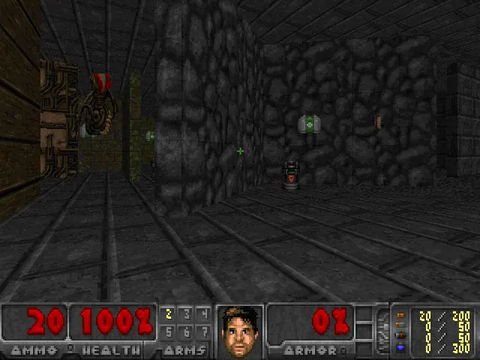}}\hfill%
  \benchmarkframe{\includegraphics{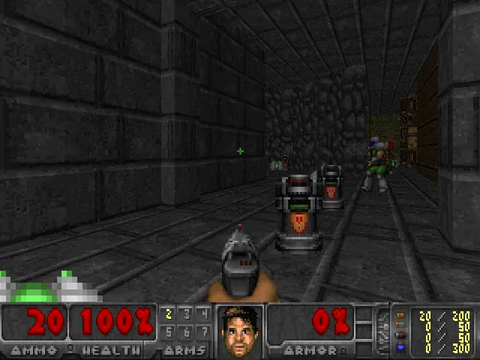}}\hfill%
  \benchmarkframe{\includegraphics{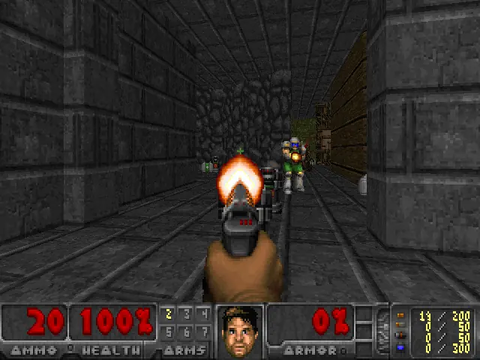}}\hfill%
  \benchmarkframe{\includegraphics{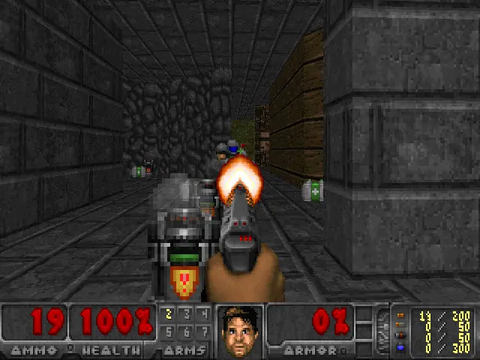}}\hfill%
  \benchmarkframe{\includegraphics{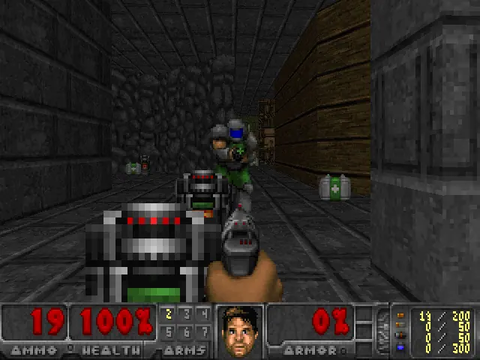}}\hfill%
  \benchmarkframe{\includegraphics{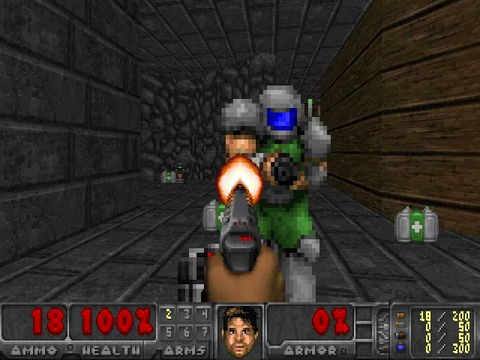}}\hfill%
  \benchmarkframe{\includegraphics{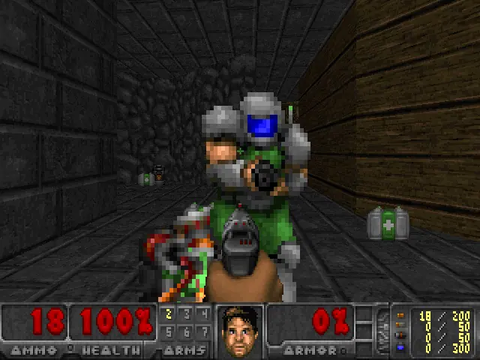}}%
}
\addvspace{4pt}
\benchmarktrajectory[``Search the damaging maze for medical cases, avoid poison bottles, and survive as long as possible.'']{Health Gathering Supreme}{%
  \benchmarkframe{\includegraphics{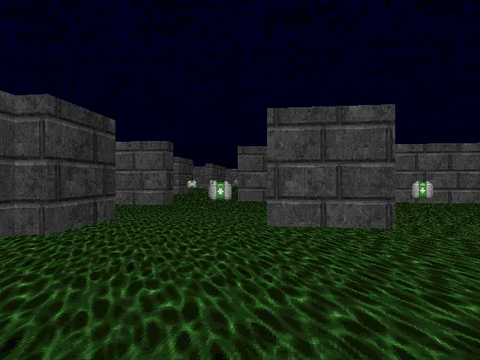}}\hfill%
  \benchmarkframe{\includegraphics{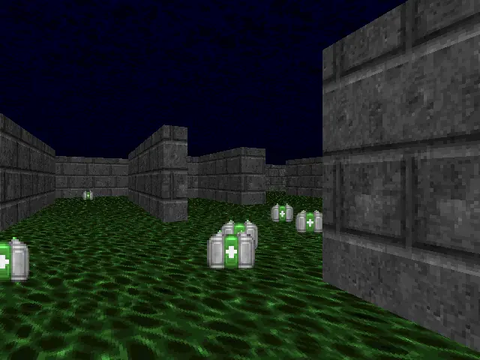}}\hfill%
  \benchmarkframe{\includegraphics{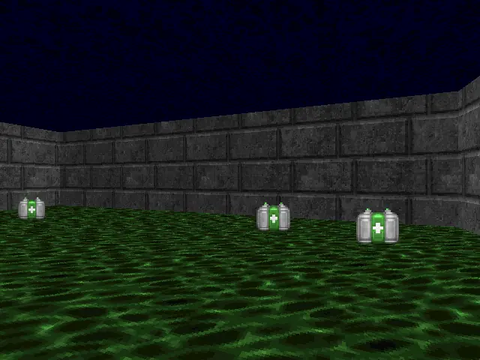}}\hfill%
  \benchmarkframe{\includegraphics{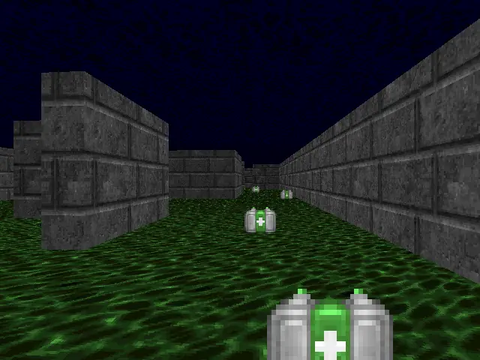}}\hfill%
  \benchmarkframe{\includegraphics{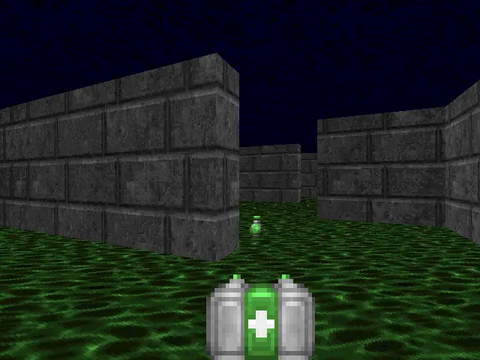}}\hfill%
  \benchmarkframe{\includegraphics{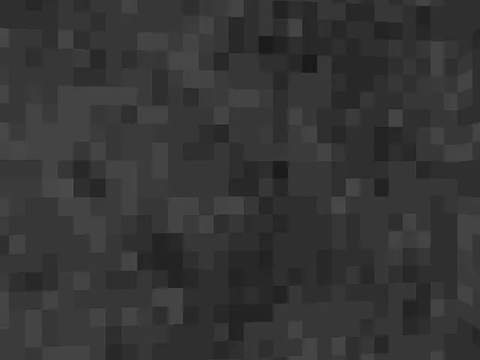}}\hfill%
  \benchmarkframe{\includegraphics{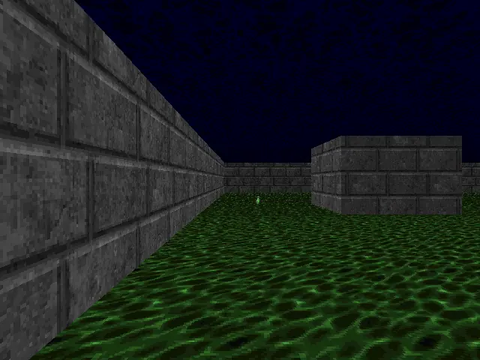}}\hfill%
  \benchmarkframe{\includegraphics{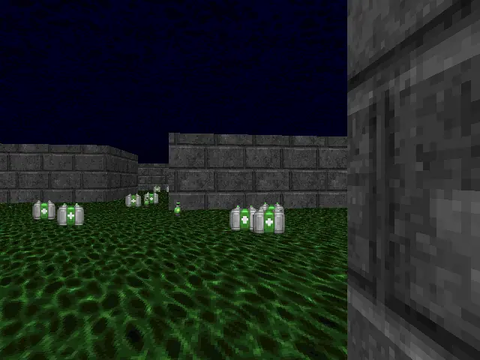}}%
}
\caption{Example tasks and scenes from ViZDoom-10.}
\label{fig:vizdoom-trajectories}
\end{figure}
\endgroup

\begingroup
\setlength{\parskip}{0pt}
\begin{figure}[tbp]
\centering
\benchmarktrajectory[``Load into level 3. Clear level 3 and reach level 4 or higher.'']{OVO}{%
  \benchmarkframe{\includegraphics{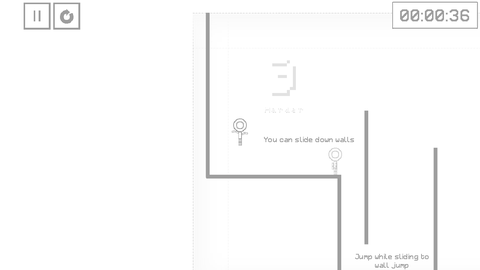}}\hfill%
  \benchmarkframe{\includegraphics{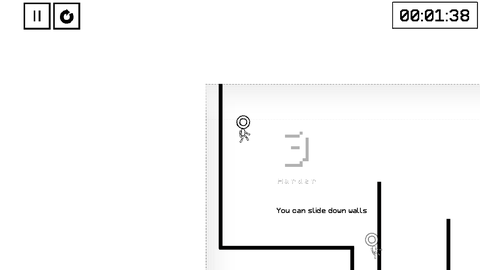}}\hfill%
  \benchmarkframe{\includegraphics{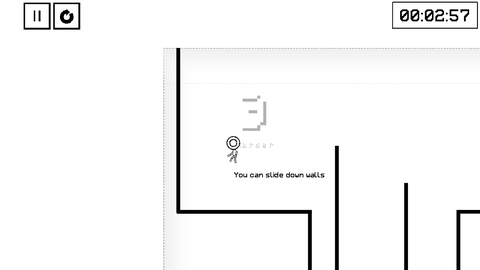}}\hfill%
  \benchmarkframe{\includegraphics{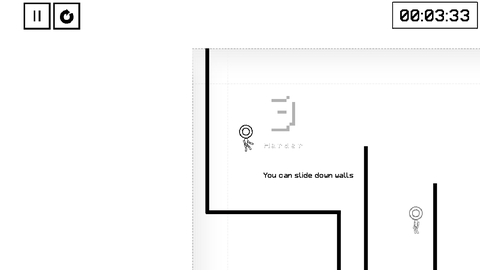}}\hfill%
  \benchmarkframe{\includegraphics{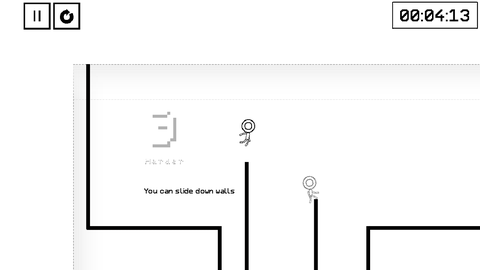}}\hfill%
  \benchmarkframe{\includegraphics{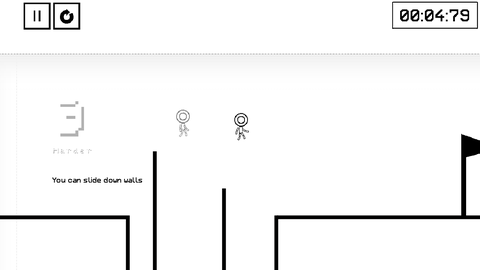}}\hfill%
  \benchmarkframe{\includegraphics{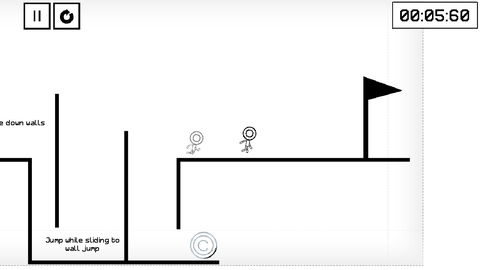}}\hfill%
  \benchmarkframe{\includegraphics{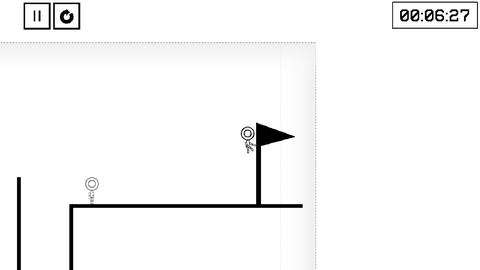}}%
}
\addvspace{4pt}
\benchmarktrajectory[``Reach at least 4000 points in one run with no stumbles, deaths, or revivals.'']{Temple Run 2}{%
  \benchmarkframe{\includegraphics{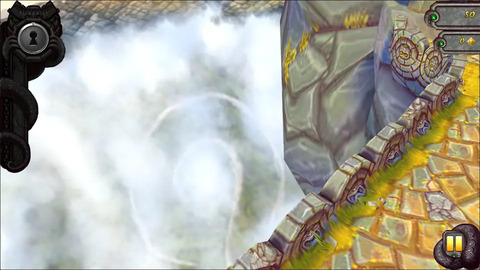}}\hfill%
  \benchmarkframe{\includegraphics{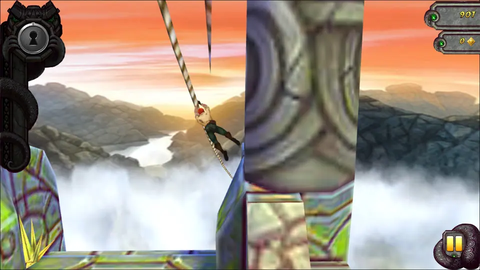}}\hfill%
  \benchmarkframe{\includegraphics{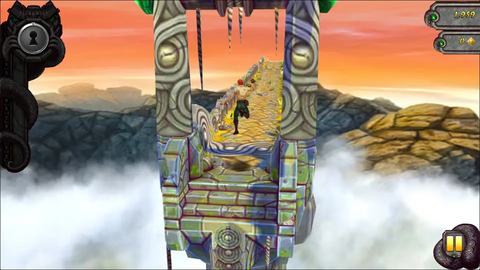}}\hfill%
  \benchmarkframe{\includegraphics{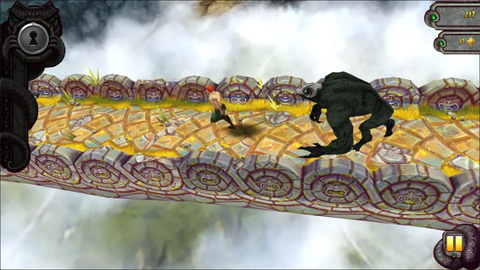}}\hfill%
  \benchmarkframe{\includegraphics{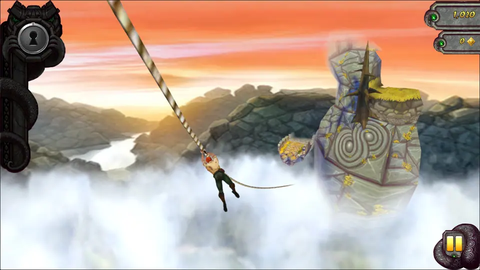}}\hfill%
  \benchmarkframe{\includegraphics{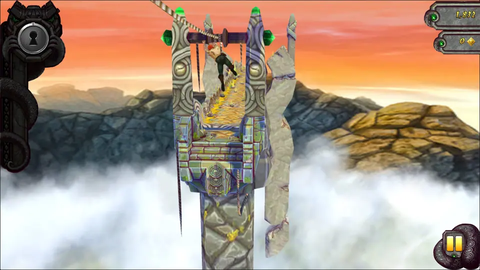}}\hfill%
  \benchmarkframe{\includegraphics{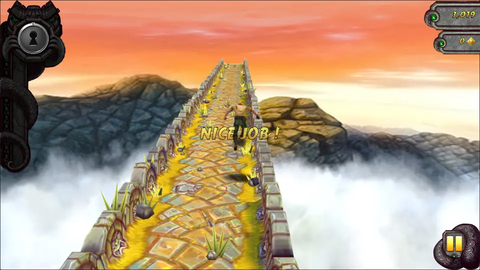}}\hfill%
  \benchmarkframe{\includegraphics{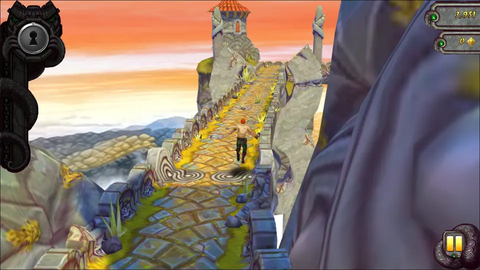}}%
}
\addvspace{4pt}
\benchmarktrajectory[``Gather at least 4 coins in the current level.'']{Another Gentleman's Adventure}{%
  \benchmarkframe{\includegraphics{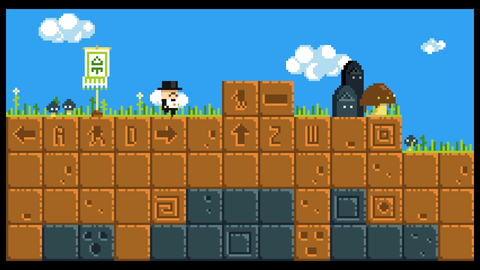}}\hfill%
  \benchmarkframe{\includegraphics{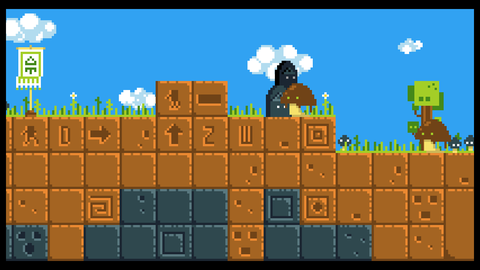}}\hfill%
  \benchmarkframe{\includegraphics{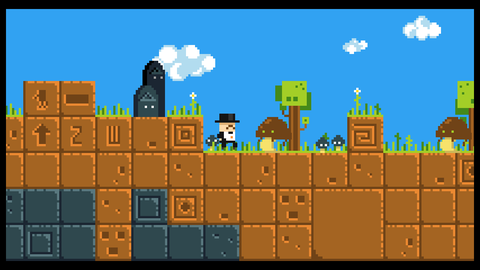}}\hfill%
  \benchmarkframe{\includegraphics{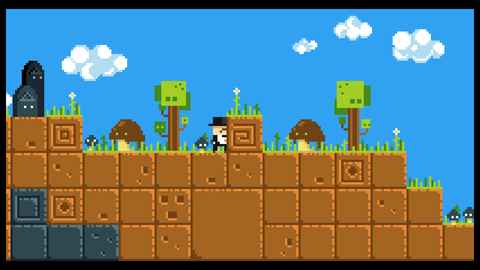}}\hfill%
  \benchmarkframe{\includegraphics{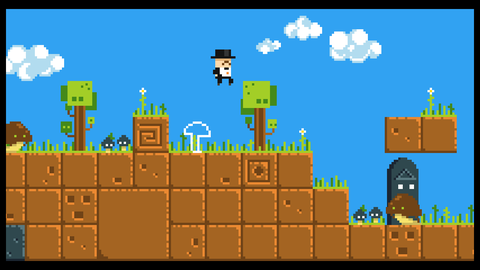}}\hfill%
  \benchmarkframe{\includegraphics{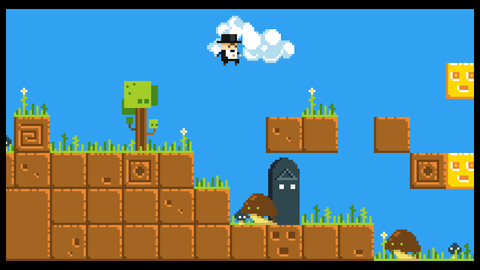}}\hfill%
  \benchmarkframe{\includegraphics{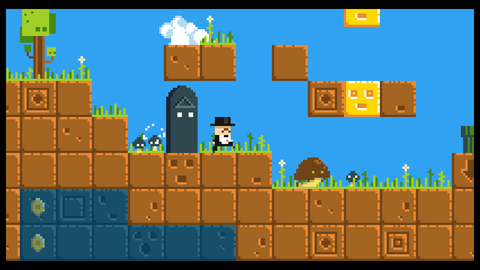}}\hfill%
  \benchmarkframe{\includegraphics{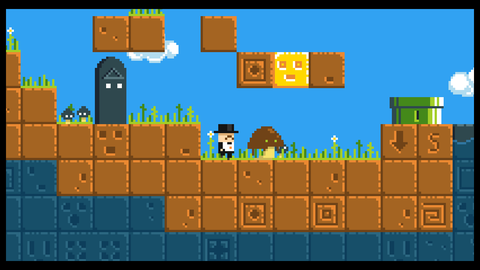}}%
}
\caption{Example tasks and scenes from GameWorld.}
\label{fig:gameworld-trajectories}
\end{figure}
\endgroup

\subsection{GameWorld}
GameWorld includes 170 tasks across 34 diverse browser games. We evaluate our model for three trials with seeds 0, 1, and 2 (Table~\ref{tab:evaluation-overview}).
The policy receives screenshots, rules, and instructions, while evaluator state is hidden.
Action duration, observation geometry, and admissible controls follow each game's configuration, and multi-role games keep separate histories.
Success is the fraction of trials reporting completion, while progress is the achieved fraction from its initial value to its target, clipped to $[0,1]$.
The evaluator averages objectives and retains the best task progress reached.
Reported success and progress average trials within each task, then equally across tasks.

\subsection{General-Purpose Capabilities}
\label{app:general-retention}

Table~\ref{tab:general-retention-panel} summarizes the evaluation panel, including the number of questions and independent answers generated per question.
Table~\ref{tab:general-retention-details} reports results on all 13 evaluation settings for the 9B and 35B-A3B model groups, complementing the domain-level summary in Table~\ref{tab:general-retention}.

All 13 evaluation settings use a maximum of 32,768 generated tokens, a total context window of 65,536 tokens, temperature 0.7, top-$p$ 0.95, and top-$k$ 20.
Each call generates one answer, with random seeds starting at 42 and increasing by one across repeated passes.
We use each checkpoint's native tokenizer and chat template with thinking enabled.
Images use a maximum pixel budget of 1,048,576.
Videos are sampled uniformly over their actual duration, with at most 256 frames and a maximum pixel budget of 196,608 per frame.

\begin{table}[H]
\centering
\caption{General-purpose evaluation panel. Repeats gives the number of independent answers generated per question. All models use the same fixed questions.}
\label{tab:general-retention-panel}
\papertablesize
\setlength{\tabcolsep}{2pt}
\begin{tabular*}{\columnwidth}{@{\extracolsep{\fill}}llrr@{}}
\toprule
\textbf{Domain} & \textbf{Benchmark} & \textbf{Questions} & \textbf{Repeats} \\
\midrule
Text & AIME 2025 & 30 & 32 \\
 & AIME 2026 & 30 & 32 \\
 & GPQA-Diamond & 198 & 2 \\
 & RG-Math & 124 & 1 \\
 & RG-Algorithmic & 144 & 1 \\
 & RG-Cognition & 96 & 1 \\
 & RG-Logic & 36 & 1 \\
\midrule
Code & LiveCodeBench v6 & 175 & 1 \\
\midrule
Image & MMMU-Pro Standard-10 & 150 & 3 \\
 & MMMU-Pro Vision & 150 & 3 \\
 & BabyVision (multiple choice) & 135 & 3 \\
 & MATH-Vision testmini & 304 & 3 \\
\midrule
Video & LVBench & 200 & 3 \\
\bottomrule
\end{tabular*}
\end{table}

\begin{table}[H]
\centering
\caption{Per-benchmark general-purpose scores (\%) underlying Table~\ref{tab:general-retention}.
Within each model group, Base denotes the Qwen3.5 backbone, SFT the supervised Player-1 checkpoint, and RL the final Player-1 checkpoint.
GS-S1 and GS-S3 denote GameScaling-9B-S1 and GameScaling-9B-S3.}
\label{tab:general-retention-details}
\papertablesize
\setlength{\tabcolsep}{2pt}
\begin{tabular*}{\linewidth}{@{\extracolsep{\fill}}lrrrrr@{\hspace{1.8em}\extracolsep{\fill}}rrr@{}}
\toprule
\multirow{2}{*}{\textbf{Benchmark}} & \multicolumn{5}{c}{\textbf{9B models}} & \multicolumn{3}{c}{\textbf{35B-A3B models}} \\
\cmidrule(lr){2-6}\cmidrule(l){7-9}
& \textbf{Base} & \textbf{GS-S1} & \textbf{GS-S3} & \textbf{SFT} & \textbf{RL}
& \textbf{Base} & \textbf{SFT} & \textbf{RL} \\
\midrule
AIME 2025 & 60.73 & 1.88 & 0.21 & 59.58 & 61.67 & 80.31 & 76.56 & 83.02 \\
AIME 2026 & 68.23 & 2.71 & 0.73 & 66.35 & 70.52 & 83.65 & 79.06 & 85.10 \\
GPQA-Diamond & 66.41 & 40.40 & 28.03 & 67.17 & 70.96 & 80.56 & 80.05 & 80.81 \\
RG-Math & 70.16 & 33.87 & 29.03 & 69.35 & 66.94 & 73.39 & 70.16 & 79.03 \\
RG-Algorithmic & 43.06 & 6.25 & 10.42 & 39.58 & 38.89 & 47.92 & 50.69 & 47.22 \\
RG-Cognition & 46.88 & 8.33 & 9.38 & 46.88 & 47.92 & 55.21 & 50.00 & 50.00 \\
RG-Logic & 86.11 & 25.00 & 52.78 & 80.56 & 77.78 & 91.67 & 83.33 & 86.11 \\
\midrule
LiveCodeBench v6 & 50.29 & 22.29 & 20.00 & 48.00 & 53.14 & 60.00 & 60.00 & 65.71 \\
\midrule
MMMU-Pro Standard-10 & 63.78 & 39.11 & 32.67 & 64.44 & 62.22 & 71.33 & 72.22 & 71.11 \\
MMMU-Pro Vision & 40.22 & 26.44 & 24.89 & 39.11 & 46.67 & 62.89 & 59.56 & 56.00 \\
BabyVision (multiple choice) & 26.42 & 15.31 & 27.16 & 23.95 & 20.49 & 32.84 & 36.54 & 34.57 \\
MATH-Vision testmini & 64.25 & 25.33 & 23.03 & 63.82 & 64.25 & 71.71 & 72.48 & 71.60 \\
\midrule
LVBench & 32.83 & 21.83 & 20.50 & 30.00 & 34.83 & 32.00 & 33.17 & 32.83 \\
\bottomrule
\end{tabular*}
\end{table}

\bibliographystyle{plainnat}
\bibliography{main}
\end{document}